\documentclass{article}

\PassOptionsToPackage{numbers}{natbib}

\newcommand{\bt}{{{\bf t}}}
\newcommand{\bs}{{{\bf s}}}

\newcommand{\bp}{{{\bf p}}}

\usepackage[final]{neurips_2025}

\usepackage[utf8]{inputenc} 
\usepackage[T1]{fontenc}    
\usepackage{hyperref}       
\usepackage{url}            
\usepackage{booktabs}       
\usepackage{amsfonts}       
\usepackage{nicefrac}       
\usepackage{microtype}      
\usepackage{multirow}

\usepackage{wrapfig}

\RequirePackage{times}    
\RequirePackage{xspace}
\RequirePackage[dvipsnames]{xcolor}
\RequirePackage{graphicx}
\RequirePackage{amsmath}
\RequirePackage{amssymb}
\RequirePackage{booktabs}
\usepackage{subcaption}

\title{Dataset Distillation of 3D Point Clouds \\ via Distribution Matching}

\author{%
  Jae-Young Yim$^\dagger$, Dongwook Kim$^\dagger$, and Jae-Young Sim$^\star$ \\
  Ulsan National Institute of Science and Technology \\
  \texttt{\{yimjae0, donguk071, jysim\}@unist.ac.kr} \\
  \\
  $^\dagger$~Equal contribution \quad $^\star$~Corresponding author
}

\begin{document}

\maketitle

\begin{abstract}
Large-scale datasets are usually required to train deep neural networks; however, they increase computational complexity, hindering practical applications. Recently, dataset distillation for images and texts has attracted considerable attention, as it reduces the original dataset to a small synthetic one to alleviate the computational burden of training while preserving essential task-relevant information.
However, dataset distillation for 3D point clouds remains largely unexplored, as point clouds exhibit fundamentally different characteristics from those of images, making this task more challenging. In this paper, we propose a distribution-matching-based distillation framework for 3D point clouds that jointly optimizes the geometric structures and orientations of synthetic 3D objects.
To address the semantic misalignment caused by the unordered nature of point clouds, we introduce a Semantically Aligned Distribution Matching (SADM) loss, which is computed on the sorted features within each channel. Moreover, to handle rotational variations, we jointly learn optimal rotation angles while updating the synthetic dataset to better align with the original feature distribution.
Extensive experiments on widely used benchmark datasets demonstrate that the proposed method consistently outperforms existing dataset distillation approaches, achieving higher accuracy and strong cross-architecture generalization.
\end{abstract}

\section{Introduction}


With the increasing demand for large-scale training datasets, the high cost of retraining models from scratch has become a major challenge, motivating research on dataset distillation~\cite{dd}. The objective of dataset distillation is to produce a significantly smaller synthetic dataset from a large original dataset, that preserves the essential task-relevant information contained in the original dataset. Hence the models trained on the reduced synthetic dataset are encouraged to achieve comparable performance to those trained on the original dataset. Existing dataset distillation methods~\cite{mtt1, mtt2, mtt3, att, gm, dm, idm, datadam, m3d, exploit, divers} can be broadly classified into gradient matching, trajectory matching, and distribution matching approaches. The gradient matching and trajectory matching methods aim to ensure that the synthetic dataset produces similar optimization dynamics to the original dataset. To this end, the gradient matching method~\cite{gm} minimizes the difference between the gradients computed on the synthetic and original datasets. However, it potentially overlooks long-term dependencies in the training process. Rather than comparing individual gradients, the trajectory matching methods~\cite{mtt1, mtt2, mtt3, att} encourage the models trained on the synthetic dataset follow similar optimization trajectories to those trained on the original dataset. Note that these methods train the networks while optimizing the synthetic dataset, and significantly increase the computational complexity. To alleviate the computational burden in dataset distillation, the distribution matching techniques~\cite{dm, idm, datadam, m3d, exploit} have been introduced. The networks are randomly initialized and used without training to extract features, after which the feature distributions of the original and synthetic datasets are compared.



While dataset distillation has been extensively studied for structured data such as images~\cite{gm, mtt1, dm} and texts~\cite{textdd1, textdd2, textdd3}, its application to 3D point clouds remains almost unexplored. PCC~\cite{pcc} simply applied an existing distillation method for images~\cite{gm} to the 3D point clouds without considering inherent characteristics of 3D point clouds, however it still suffers from high computational complexity. Unlike the structured data, 3D point clouds consist of unordered and irregularly distributed points in 3D space. Therefore, semantically similar regions across different 3D models are often associated with inconsistent orders (i.e., point indices), which makes direct feature comparison incorrect, thereby worsening the performance of distribution-matching-based dataset distillation. We refer to this issue as {\it semantic misalignment}.
Moreover, the datasets of 3D point clouds also suffers from {\it rotational variation}. 3D point clouds are often captured or synthesized under arbitrary poses due to the absence of canonical orientations. Such rotational differences cause objects of the same class to produce different features, significantly increasing intra-class variability. As a result, it becomes difficult to construct a representative synthetic dataset that faithfully follows the feature distribution of the original dataset.

In this paper, we propose an optimization framework combining Semantically Aligned Distribution Matching (SADM) and orientation optimization for dataset distillation of 3D point clouds. A major challenge in this setting is achieving effective feature alignment despite the unordered nature of point clouds and their arbitrary orientations. The SADM loss addresses the issue of misaligned semantic structures by sorting the point-wise feature values within each channel before computing the distance between two feature distributions. In parallel, we employ learnable rotation parameters to address the orientation variation by estimating the optimal poses of 3D models while generating the synthetic dataset. By simultaneously enforcing semantic consistency and orientation alignment, the proposed method achieves superior performance of dataset distillation compared with the existing methods, as demonstrated by extensive experiments on standard 3D point cloud classification benchmarks.

The key contributions of this paper are summarized as follows:
\begin{itemize}
    \item To the best of our knowledge, we are the first to propose a distribution-matching-based dataset distillation method of 3D point clouds, that jointly optimizes the shapes and orientations of synthetic dataset. 
    \item We devised the SADM loss, computed on the sorted features within each channel, to preserve semantic alignment between the compared 3D objects.
    \item We validated the proposed method through extensive experiments on the four benchmark datasets widely used for 3D point clouds classification, and showed the superiority of the proposed method over the existing dataset distillation techniques.
\end{itemize}

\section{Related Works}

\subsection{3D Point Data Analysis}
3D point clouds are generally unordered exhibiting irregular characteristics, that makes it difficult to apply the convolution operations in deep neural networks commonly used for images. To process such data, early studies~\cite{(multiview)multi, (voxel)voxelnet} converted point clouds into structured representations, such as multi-view images or voxel grids, enabling the use of standard convolutional neural networks (CNN) architectures. However, these conversions introduce quantization errors and increase memory overhead. PointNet~\cite{pointnet} addressed this limitation by directly learning the features from unordered points, which serves as the backbone in our dataset distillation framework. Subsequent methods, such as PointNet++~\cite{pointnet2}, PointConv~\cite{pointconv}, and DGCNN~\cite{dgcnn}, extended PointNet by capturing local geometric relationships through hierarchical architecture. Additionally, attempts have been also made to apply the transformer~\cite{transformer} to 3D point clouds processing, where Point Transformer~\cite{pointtransformer} utilizes the attention mechanism to capture the long-range dependencies. We adopt these architectures as evaluation networks to assess the cross-architecture generalization capability of our distilled datasets.

\subsection{Coreset Selection}
Coreset selection is a technique designed to select representative samples from a given dataset, while maintaining the model's performance even with the sampled data. The random selection method~\cite{herding(icarl)} randomly chooses a subset of data samples from the whole dataset. It is simple but suffers from the robustness due to the lack of informative criteria for sampling. The K-center method~\cite{kcenter} selects data samples iteratively that maximize the minimum distance to the set of already selected ones, taking into account the data distribution. The Herding method~\cite{herding(e2e),herding(scail)} iteratively takes data points that minimize the discrepancy between the mean embeddings of the selected subset and the entire dataset in the feature space. The coreset selection operates within the space of existing samples, and hence lacks the flexibility to synthesize informative patterns. This limitation naturally leads to the emergence of dataset distillation, which generates synthetic datasets capturing task-specific patterns beyond the original samples.

\subsection{Dataset Distillation}
Dataset distillation~\cite{dd} methods can be largely categorized into the gradient matching~\cite{gm}, trajectory matching~\cite{mtt1, mtt2, mtt3, att}, and distribution matching~\cite{dm, idm, datadam, m3d, exploit} approaches. The gradient matching, first introduced by DC~\cite{gm}, minimizes the difference between the gradients computed from the original and synthetic datasets, respectively, guiding the synthetic dataset to follow the training direction of the original dataset. The trajectory matching methods~\cite{mtt1, mtt2, mtt3,att}, initially proposed by MTT~\cite{mtt1}, encourage the models trained on the synthetic dataset to follow the optimization trajectories similar to those trained on the original dataset rather than comparing individual gradients. ATT~\cite{att} automatically adjusts the length of the training trajectories between the synthetic and original datasets, enabling more effective and precise matching. The distribution matching, introduced by DM~\cite{dm}, focuses on minimizing the distance between the feature distributions of the original and synthetic datasets. DataDAM~\cite{datadam} enhances the distribution matching by aligning the feature maps using attention mechanism, achieving unbiased feature representation with low computational overhead. Furthermore, M3D~\cite{m3d} employs the Gaussian kernel function in the distribution matching loss, enabling the alignment across higher-order statistical characteristics of the feature distributions. Recently, a gradient matching-based point clouds distillation method~\cite{pcc} has been introduced, however it simply applied the existing image-based method without considering the unique characteristics of 3D point clouds.

\section{Methodology}

\begin{figure*}[t]
	\centering
	\includegraphics[width=1\linewidth]{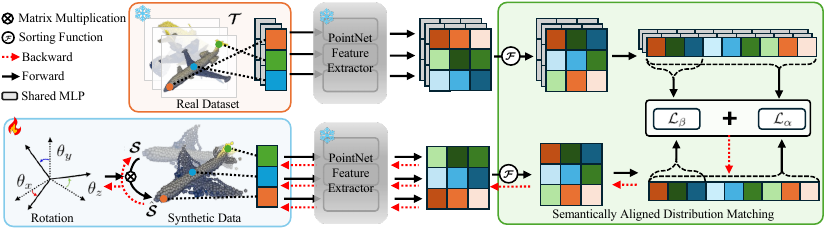}
	\caption{The overall framework of the proposed dataset distillation method for 3D point clouds.}
	\label{fig:fig1} \vspace{-2mm}
\end{figure*}

We propose a novel dataset distillation method for 3D point clouds that first addresses the key challenges of semantic misalignment and rotational variation. We perform SADM that effectively aligns the features of semantically consistent structures across different 3D models. To handle the rotational variation, we also estimate optimal orientations while generating the synthetic dataset. Figure~\ref{fig:fig1} shows the overall framework of the proposed method. 

\subsection{Preliminaries}

{\bf Problem Definition. }The objective of dataset distillation is to compress the task-relevant information in the original dataset ${\boldsymbol{\mathcal T}}=\left\{ \bt_i \right\}_{i=1}^{|\mathcal{T}|}$ and generate a much smaller synthetic dataset ${\boldsymbol{\mathcal S}}=\left\{\bs_i\right\}_{i=1}^{|\mathcal{S}|}$, where $|{\boldsymbol{\mathcal S}}|\ll|{\boldsymbol{\mathcal T}}|$, such that a model trained on ${\boldsymbol{\mathcal S}}$ achieves close performance to that trained on ${\boldsymbol{\mathcal T}}$. Given a 3D point cloud sample $\bp$ following a real data distribution with the corresponding class label $l$, the optimal synthetic dataset ${\boldsymbol{\mathcal S}}^{\star}$ can be obtained via 
\begin{equation} \label{eq:eq1}
	{\boldsymbol{\mathcal S}}^{\star} = \mathop{\arg \min}\limits_{{\boldsymbol{\mathcal S}}} \mathbb{E}_{\bp} \left[||{\mathcal L}(\phi_{{{\boldsymbol{\mathcal T}}}}(\bp),l)-{\mathcal L}(\phi_{{{\boldsymbol{\mathcal S}}}}(\bp),l)||^2 \right],
\end{equation}
where ${\mathcal L}$ denotes a task-specific loss function, such as the cross-entropy loss, and $\phi_{{\boldsymbol{\mathcal T}}}$ and $\phi_{{\boldsymbol{\mathcal S}}}$ represent the models to estimate the class label which are trained on ${\boldsymbol{\mathcal T}}$ and ${\boldsymbol{\mathcal S}}$, respectively.

{\bf Distribution Matching.} The distribution matching (DM) strategy focuses on aligning the feature distributions derived from the original and synthetic datasets, respectively, via
\begin{equation} \label{eq:eq2}
    {\boldsymbol{\mathcal S}}^\star = \mathop{\arg\min}_{{\boldsymbol{\mathcal S}}} D(\phi({\boldsymbol{\mathcal T}}), \phi({\boldsymbol{\mathcal S}})),
\end{equation}
where $\phi$ is the feature extractor and $D$ denotes a distance function. We employ a randomly initialized, untrained network, which has been demonstrated to sufficiently capture the structural information for distribution alignment~\cite{dm}. 
Also, the Maximum Mean Discrepancy~(MMD)~\cite{mmd} loss $\mathcal L_{\text{MMD}}$ is often used as $D$, given by
\begin{equation} \label{eq:eq3}
    \mathcal{L}_{\text{MMD}}({\boldsymbol{\mathcal T}}, {\boldsymbol{\mathcal S}}) = {K}({\boldsymbol{\mathcal T}}, {\boldsymbol{\mathcal T}}) + {K}({\boldsymbol{\mathcal S}}, {\boldsymbol{\mathcal S}}) - 2{K}({\boldsymbol{\mathcal T}}, {\boldsymbol{\mathcal S}}),
\end{equation}
where ${K}(\cdot, \cdot)$ is a kernel function. We used the Gaussian kernel in this work as
\begin{equation} \label{eq:eq4}
    {K}({\boldsymbol{\mathcal T}}, {\boldsymbol{\mathcal S}}) = \frac{1}{|\boldsymbol{\mathcal T}| \cdot |\boldsymbol{\mathcal S}|} \sum_{\bt \in \boldsymbol{\mathcal T}} \sum_{\bs \in \boldsymbol{\mathcal S}} \exp\left( -\frac{\|\phi(\bt) - \phi(\bs)\|^2}{2\sigma} \right).
\end{equation}

\subsection{Semantically Aligned Distribution Matching}
\begin{wrapfigure}{r}{0.5\linewidth}
	\centering
	\includegraphics[width=1\linewidth]{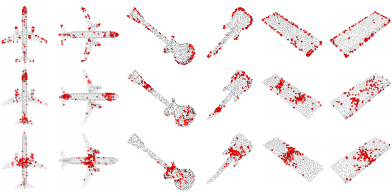}
	\caption{Visualization of the points corresponding to the largest (top), 200th largest (middle), and 500th largest (bottom) features in each channel.}
	\label{fig:fig2}
\end{wrapfigure}

Whereas the pixels of image exhibit well structured spatial relationships with one another and consistently indexed across different images, the points in 3D point clouds are unordered with different indices across different models. Therefore, the conventional distribution matching methods cannot be directly applied to 3D point clouds, since the features of semantically similar structures are not aligned between two compared models. Motivated by the importance of preserving semantic correspondence, we investigate the relationship between the feature values and semantic significance in 3D point clouds. We first extracted point-wise features using a randomly initialized network $\phi$, and sorted the feature values for each of 1024 channels according to their size. Figure~\ref{fig:fig2} visualizes the points corresponding to the largest (top), 200th largest (middle), and 500th largest (bottom) features, respectively. We observe that, even without model training, the points associated with the largest features consistently capture semantically meaningful regions, such as edges and corners, across different classes. In contrast, the points yielding low-ranked features are distributed around less informative regions. This suggests that the features reflect the relative importance of points in characterizing the structures of 3D objects, and the points with similar orders of sorted features tend to capture semantically related regions across different 3D models.

In order to preserve the semantic correspondence across different 3D models, we propose a Semantically Aligned Distribution Matching (SADM) loss. Specifically, given a 3D point cloud object $\bp$, we extract the features using $\phi$ via
\begin{equation} \label{eq:eq5}
    \phi(\bp) =
    \begin{bmatrix}
    f_{1,1} & f_{2,1} & \cdots & f_{C,1} \\
    f_{1,2} & f_{2,2} & \cdots & f_{C,2} \\
    \vdots  & \vdots  & \ddots & \vdots  \\
    f_{1,N} & f_{2,N} & \cdots & f_{C,N}
    \end{bmatrix}
    = \left[{\bf f}_1, {\bf f}_2, \dots, {\bf f}_C\right], \quad {\bf f}_i \in \mathbb{R}^{N},
\end{equation}
where $N$ denotes the number of points in $\bp$, $C$ is the number of feature channels, and ${\bf f}_i=\left\{f_{i,1}, f_{i,2}, ..., f_{i,N}\right\}$ represents the feature vector of the \(i\)-th channel where $f_{i,j}$ is the feature value of the $j$-th point. We then perform channel-wise sorting to the extracted features of $f_{i,j}$'s in the descending order of their values, and obtain the sorted feature vector $\tilde{{\bf f}}_i=\left\{\tilde{f}_{i,1}, \tilde{f}_{i,2}, ..., \tilde{f}_{i,N}\right\}$ such that $\tilde{f}_{i,j} \geq \tilde{f}_{i,j+1}$. Then we have the set of the sorted feature vectors as
\begin{equation} \label{eq:eq6}
	\tilde{\phi}(\bp) = \left[\tilde{{\bf f}}_1, \tilde{{\bf f}}_2, \dots, \tilde{{\bf f}}_C\right], \quad \tilde{{\bf f}}_i \in \mathbb{R}^{N}.
\end{equation}

Note that $\bt$ in the original dataset $\boldsymbol{\mathcal T}$ and $\bs$ in the synthetic dataset $\boldsymbol{\mathcal S}$ exhibit different orders of points in general, and therefore the features in ${\phi}(\bt)$ and ${\phi}(\bs)$ are inconsistently aligned with each other worsening the desired behavior of comparison in~\eqref{eq:eq3}. On the contrary, the sorted features $\tilde{\phi}(\bt)$ and $\tilde{\phi}(\bs)$ exhibit consistent ordering of features in each channel reflecting their relative semantic importance, and thus facilitates reliable feature comparison across different 3D objects. To measure the discrepancy between the sorted features, we redefine the Gaussian kernel as
\begin{equation} \label{eq:eq7}
\tilde{K}({{\boldsymbol{\mathcal T}}, \boldsymbol{\mathcal S}})=\frac{1}{|\boldsymbol{\mathcal T}| \cdot |\boldsymbol{\mathcal S}|} \sum_{{\bt \in \boldsymbol{\mathcal T}}} \sum_{\bs \in \boldsymbol{\mathcal S}} \exp\left(-\frac{||\tilde{\phi}(\bt)- \tilde{\phi}(\bs)||^2}{2\sigma}\right).
\end{equation}
Then we devise the loss $\mathcal{L}_{\alpha}$ from $\mathcal{L}_{\text{MMD}}$~\eqref{eq:eq3}, given by
\begin{equation}
\mathcal{L}_{\alpha}(\boldsymbol{\mathcal{T}}, \boldsymbol{\mathcal{S}}) = \tilde{K}(\boldsymbol{\mathcal{T}}, \boldsymbol{\mathcal{T}}) + \tilde{K}(\boldsymbol{\mathcal{S}}, \boldsymbol{\mathcal{S}}) - 2\tilde{K}(\boldsymbol{\mathcal{T}}, \boldsymbol{\mathcal{S}}).
\end{equation}
We additionally employ $\mathcal{L}_{\beta}$ to boost the role of the most significant feature in each channel, defined as
\begin{equation}
\mathcal{L}_{\beta}(\boldsymbol{\mathcal{T}}, \boldsymbol{\mathcal{S}}) = \tilde{K}_{\text{top}}(\boldsymbol{\mathcal{T}}, \boldsymbol{\mathcal{T}}) + \tilde{K}_{\text{top}}(\boldsymbol{\mathcal{S}}, \boldsymbol{\mathcal{S}}) - 2 \tilde{K}_{\text{top}}(\boldsymbol{\mathcal{T}}, \boldsymbol{\mathcal{S}}),
\end{equation}
where $\tilde{K}_{\text{top}}(\cdot, \cdot)$ denotes the Gaussian kernel computed with only the largest feature in each channel.
The SADM loss is then formulated as a weighted sum of the two losses, given by
\begin{equation} \label{eq:eq8}
\mathcal{L}_{\text{SADM}}(\boldsymbol{\mathcal{T}}, \boldsymbol{\mathcal{S}}) = \lambda_1 \mathcal{L}_{\alpha}(\boldsymbol{\mathcal{T}}, \boldsymbol{\mathcal{S}}) + \lambda_2 \mathcal{L}_{\beta}(\boldsymbol{\mathcal{T}}, \boldsymbol{\mathcal{S}}),
\end{equation}
where $\lambda_1, \lambda_2$ are the weighting parameters. By using $\mathcal{L}_{\text{SADM}}$ instead of $\mathcal{L}_{\text{MMD}}$, we facilitate reliable matching of feature distributions considering semantic structures of 3D point clouds, thereby improving the performance dataset distillation.

\subsection{Estimation of Optimal Rotations}
3D objects usually exhibit different orientations from each other. Therefore, while generating optimal 3D objects in the synthetic dataset in terms of their geometric shapes, we also estimate their optimal rotations best representing various orientations of 3D objects in the original dataset.
In practice, we introduce three rotation angles, $\theta_x$, $\theta_y$, and $\theta_z$, corresponding to rotations around the $x$-, $y$-, and $z$-axes, respectively. These angles are treated as learnable parameters, allowing the orientation of each synthetic 3D object in ${\boldsymbol{\mathcal{S}}}$ to be adaptively adjusted during dataset distillation. Therefore, instead of minimizing $\mathcal{L}_{\text{SADM}}({\boldsymbol{\mathcal{T}}, \boldsymbol{\mathcal{S}}})$ in~\eqref{eq:eq8}, we minimize $\mathcal{L}_{\text{SADM}}({\boldsymbol{\mathcal{T}}, \mathcal{R}_{\boldsymbol{\theta}}(\boldsymbol{\mathcal{S}})})$, where $\mathcal{R}_{\boldsymbol{\theta}}$ denotes the rotation operator according to the rotation parameters $\boldsymbol{\theta} = (\theta_x, \theta_y, \theta_z)$. The overall objective is formally defined as
\begin{equation}
	\left\{{\boldsymbol{\mathcal S}}^\star, {\boldsymbol{\theta}}^\star \right\}= \mathop{\arg\min}_{\left\{{\boldsymbol{\mathcal S}}, {\boldsymbol{\theta}}\right\}} \mathcal{L}_{\text{SADM}}({\boldsymbol{\mathcal T}}, \mathcal{R}_{\boldsymbol{\theta}}(\boldsymbol{\mathcal{S}})).
\end{equation}

Note that we optimize the synthetic 3D objects as well as their rotation parameters simultaneously, during the dataset distillation process. Specifically, at each iteration, we randomly initialize the network parameters and construct the synthetic dataset $\boldsymbol{\mathcal{S}}$ by randomly selecting samples from the original dataset. We then form $\mathcal{R}_{\boldsymbol{\theta}}(\boldsymbol{\mathcal{S}})$ by rotating the objects in $\boldsymbol{\mathcal{S}}$ according to the angles $\theta_x$, $\theta_y$, and $\theta_z$, initially set to zero. For each class, a mini-batch is sampled from $\boldsymbol{\mathcal{T}}$, with a batch size of 8 per class. The corresponding synthetic mini-batch is sampled from $\mathcal{R}_{\boldsymbol{\theta}}(\boldsymbol{\mathcal{S}})$, with the batch size determined by the number of point cloud objects per class (PPC). The joint optimization process iteratively updates both the geometric structure of the synthetic dataset and their rotation parameters by minimizing $\mathcal{L}_{\text{SADM}}({\boldsymbol{\mathcal{T}}, \mathcal{R}_{\boldsymbol{\theta}}(\boldsymbol{\mathcal{S}})})$. This ensures that the synthetic dataset preserves the geometric characteristics of the original dataset while aligning their orientations more effectively. The benefit of this joint optimization is justified in the following proposition.

\textbf{Proposition 1.}
\textit{Jointly optimizing the synthetic dataset $\boldsymbol{\mathcal{S}}$ and the rotation parameters $\boldsymbol{\theta}$ guarantees a lower or equal loss to that of optimizing $\boldsymbol{\mathcal{S}}$ alone.}
\begin{equation}
\min_{\left\{\boldsymbol{\mathcal{S}}, \boldsymbol{\theta}\right\}} \mathcal{L}_{\text{SADM}}(\boldsymbol{\mathcal{T}}, \mathcal{R}_{\boldsymbol{\theta}}(\boldsymbol{\mathcal{S}})) \leq \min_{\boldsymbol{\mathcal{S}}} \mathcal{L}_{\text{SADM}}(\boldsymbol{\mathcal{T}}, \boldsymbol{\mathcal{S}}),
\end{equation}
\textit{where $\mathcal{R}_{\boldsymbol{\theta}}$ denotes the rotation operator according to $\boldsymbol{\theta}$.}

{\bf \textit{Proof Sketch.}} The proof is provided in Appendix A. Jointly optimizing over $\boldsymbol{\mathcal{S}}$ and $\boldsymbol{\theta}$ enlarges the feasible set, as the original optimization over $\boldsymbol{\mathcal{S}}$ alone is a special case of the joint optimization with fixed $(\theta_x,\theta_y,\theta_z)=(0,0,0)$. \hfill $\square$

\begin{table*}[t]
    \centering
    \caption{Comparison of quantitative performance. DC, DM, and the proposed method were initialized with random selection~\cite{herding(icarl)} for fair comparison. `Whole' refers to the classification accuracy obtained by training the network on the entire original dataset without any distillation. `Ratio' represents the percentage of the size of the distilled dataset compared to that of the original dataset. The best and the second best scores are highlighted in bold and underlined, respectively.}
    
    \resizebox{\textwidth}{!}{%
    \begin{tabular}{@{}c|ccc|ccc|ccc|ccc@{}}
    \toprule
    \textbf{Datasets} & \multicolumn{3}{c|}{\textbf{ModelNet10}~\cite{modelnet40}} & \multicolumn{3}{c|}{\textbf{ModelNet40}~\cite{modelnet40}} & \multicolumn{3}{c|}{\textbf{ShapeNet}~\cite{shapenet}} & \multicolumn{3}{c}{\textbf{ScanObjectNN}~\cite{scanobjectnn}} \\ 
    \midrule
    \textbf{PPC} & 1 & 3 & 10 & 1 & 3 & 10 & 1 & 3 & 10 & 1 & 3 & 10 \\
    \textbf{Ratio (\%)} & 0.25 & 0.75 & 2.5 & 0.4 & 1.2 & 4.0 & 0.15 & 0.45 & 1.5 & 0.15 & 0.45 & 1.5 \\ \midrule
    \textbf{Whole} & \multicolumn{3}{c|}{91.41} & \multicolumn{3}{c|}{87.84} & \multicolumn{3}{c|}{82.49} & \multicolumn{3}{c}{63.84} \\ 
    \midrule
    \midrule
    \textbf{Random}   & 35.5$\pm4.7$ & 75.2$\pm1.7$ & 85.3$\pm1.1$  & 34.6$\pm1.8$ & 60.0$\pm1.32$ & 74.1$\pm0.4$  & 34.0$\pm3.0$ & 54.8$\pm1.5$ & 63.1$\pm1.0$  & 13.9$\pm1.4$ & 20.4$\pm1.3$ & 34.8$\pm1.1$ \\ 
    
    \textbf{Herding}  & 40.1 $\pm 5.2$ & \underline{78.0} $\pm 1.3$ & \underline{86.9}$\pm 0.6$  & \underline{54.4} $\pm 2.0$ & \underline{68.5} $\pm 1.0$ & \underline{78.8} $\pm 0.4$ & 49.5 $\pm 2.3$ & 59.8 $\pm 0.9$ & \underline{66.9} $\pm 0.5$ & 15.7 $\pm 1.7$ & \underline{27.7} $\pm 1.4$ & \underline{38.7} $\pm 1.7$ \\
    
    \textbf{K-center} & 40.1 $\pm 5.2$ & 77.6$\pm 1.9$ & 83.2$\pm 1.4$ & 54.4$\pm 2.0$ & 63.0 $\pm 2.7$ & 65.3 $\pm 1.1$ & 49.5 $\pm 2.3$ & 51.4 $\pm 1.7$ & 47.8 $\pm 0.7$ & 15.7 $\pm 1.7$ & 19.8 $\pm 0.8$ & 24.0 $\pm 1.0$ \\ 
    
    \midrule
    
    \textbf{DM}   & 31.9$\pm4.3$& 77.6$\pm1.6$& 86.1$\pm0.8$& 32.3$\pm4.8$& 62.0$\pm1.9$& 75.2$\pm0.5$& 27.8$\pm4.0$& 56.1$\pm1.2$& 64.3$\pm0.7$& 14.9$\pm2.0$& 23.3$\pm1.5$& 35.6$\pm1.4$\\ 
    
    \textbf{DC}  & \underline{43.9}$\pm5.5$& 75.0$\pm2.4$& 86.1$\pm1.1$& 52.0$\pm1.7$& 66.6$\pm1.3$& 75.6$\pm0.5$& \underline{49.8}$\pm1.3$&\underline{60.1}$\pm1.4$& 64.6$\pm0.7$& \textbf{18.6}$\pm\textbf{1.6}$& 24.3$\pm2.5$& 35.3$\pm1.2$\\     
    
    \midrule
    \textbf{Ours} &  \textbf{44.7} $\pm \textbf{6.1}$&  \textbf{84.4} $\pm \textbf{1.2}$ & \textbf{87.8} $\pm \textbf{1.0}$ & \textbf{55.8} $\pm \textbf{1.5}$ & \textbf{72.1} $\pm \textbf{0.8}$ & \textbf{80.1} $\pm \textbf{0.4}$ & \textbf{50.2} $\pm \textbf{1.4}$ & \textbf{63.7} $\pm \textbf{0.7}$ & \textbf{68.4} $\pm \textbf{0.5}$ & \underline{17.4} $\pm 1.5$  & \textbf{31.6} $\pm \textbf{1.0}$  & \textbf{43.9} $\pm \textbf{1.9}$ \\
    \bottomrule
    \end{tabular}
    }
    \label{tab:tab1}
\end{table*}

\begin{table*}[t]
    \centering
    \caption{Comparison of cross-architecture generalization performance evaluated on PointNet++~\cite{pointnet2}, DGCNN~\cite{dgcnn}, PointConv~\cite{pointconv}, and PT~\cite{pointtransformer}, respectively.}
    \resizebox{\textwidth}{!}{%
    \begin{tabular}{@{}c|ccc|ccc|ccc|ccc@{}}
    \toprule
    \textbf{Datasets} & \multicolumn{3}{c|}{\textbf{ModelNet10}~\cite{modelnet40}} & \multicolumn{3}{c|}{\textbf{ModelNet40}~\cite{modelnet40}} & \multicolumn{3}{c|}{\textbf{ShapeNet}~\cite{shapenet}} & \multicolumn{3}{c}{\textbf{ScanObjectNN}~\cite{scanobjectnn}} \\ 
    \midrule
    \textbf{Method}       & \textbf{DM} & \textbf{DC} & \textbf{Ours} & \textbf{DM} & \textbf{DC} & \textbf{Ours} & \textbf{DM} & \textbf{DC} & \textbf{Ours} & \textbf{DM} & \textbf{DC} & \textbf{Ours} \\
    \midrule
    \midrule
    \textbf{PointNet++}   & 35.0$\pm8.3$ & \underline{51.5}$\pm7.8$ & \textbf{82.9}$\pm \textbf{1.4}$ & 14.1$\pm2.7$ & \underline{46.9}$\pm6.0$ & \textbf{68.5}$\pm \textbf{0.5}$ & 20.3$\pm3.7$ & \underline{34.3}$\pm3.4$ & \textbf{53.2}$\pm \textbf{1.5}$ & \underline{15.5}$\pm3.6$ & 11.6$\pm2.2$ & \textbf{22.0}$\pm\textbf{3.9}$ \\
    
    \textbf{DGCNN}        & 56.6$\pm5.3$ & \underline{62.6}$\pm2.2$ & \textbf{79.0}$\pm\textbf{2.2}$ & 34.8$\pm3.5$ & \underline{52.3}$\pm2.7$ & \textbf{66.9}$\pm\textbf{1.1}$ & 13.8$\pm2.2$ & \underline{38.0}$\pm2.7$ & \textbf{52.8}$\pm\textbf{1.6}$ & \underline{17.7}$\pm2.4$ & 12.5$\pm1.3$ & \textbf{19.0}$\pm\textbf{1.5}$ \\

    \textbf{PointConv}    & 33.4$\pm6.2$ & \underline{40.0}$\pm10.5$ & \textbf{56.6}$\pm\textbf{4.9}$ & 20.0$\pm5.7$ & \underline{37.9}$\pm3.9$ & \textbf{51.3}$\pm\textbf{6.2}$ & 17.4$\pm3.0$ & \underline{34.0}$\pm4.8$ & \textbf{47.4}$\pm\textbf{2.3}$ & \underline{15.2}$\pm1.7$ & 14.2$\pm1.8$ & \textbf{16.8}$\pm\textbf{1.5}$ \\
    
    \textbf{PT}         & 48.5$\pm6.6$ & \underline{61.8}$\pm1.6$ & \textbf{77.6}$\pm\textbf{1.6}$ & 26.7$\pm5.0$ & \underline{47.5}$\pm2.7$ & \textbf{61.6}$\pm\textbf{1.1}$ & 34.2$\pm5.3$ & \underline{42.6}$\pm2.5$ & \textbf{55.5}$\pm\textbf{0.5}$ & \underline{17.1}$\pm0.8$ & 14.9$\pm0.5$ & \textbf{21.9}$\pm\textbf{2.7}$\\
    
    \bottomrule
    \end{tabular}
    }
    \label{tab:tab2}
\end{table*}

\section{Experimental Results}
\subsection{Experimental Setup}
The proposed method was evaluated on the ModelNet10~\cite{modelnet40}, ModelNet40~\cite{modelnet40}, ShapeNet~\cite{shapenet}, and ScanObjectNN~\cite{scanobjectnn} datasets. The ModelNet10, ModelNet40, and ShapeNet are synthetic datasets generated from CAD models, containing 10, 40, and 55 classes, respectively. The ScanObjectNN consists of real-world scanned objects from 15 classes. Following previous methods~\cite{gm, dm}, we measure the classification accuracy trained on the distilled synthetic dataset. We ensure the fairness by training the network 10 times and report the mean and standard deviation of accuracy. We evaluate the performance across different PPC values of 1, 3, and 10. Each point cloud object contains 1,024 points, which is a common standard used in the 3D point clouds classification tasks~\cite{pointnet, pointnet2}. The dataset distillation process for the synthetic dataset was optimized for 1,500 iterations, where the performance was evaluated at every 250 iterations by training the PointNet on the synthetic dataset and testing it on the original test dataset. 


\subsection{Performance Comparison}
\subsubsection{Dataset Distillation}
We compare the performance of the proposed method with that of 1) the three most representative coreset selection methods: random selection~\cite{herding(icarl)}, Herding~\cite{herding(e2e), herding(scail)}, and K-Center~\cite{kcenter}, 2) two existing image dataset distillation methods of DC~\cite{gm} and DM~\cite{dm}.
Table~\ref{tab:tab1} compares the quantitative performance of the proposed method with that of the existing methods. DM and DC were initialized with random selection. The synthetic datasets were optimized using PointNet~\cite{pointnet}, and classification performance was evaluated on PointNet to demonstrate their validity. Among the coreset selection methods, Herding achieves the best overall performance. When PPC is set to 1, Herding and K-Center yield identical results since K-Center selects the first data point using the same algorithm to Herding. As PPC increases, Herding consistently outperforms other coreset selection methods, showing its effectiveness. However, because coreset selection methods rely solely on sample selection without further optimization, they struggle to capture complex feature distributions.
The features used in DM are extracted from the layer before the classifier, and they discard substantial information through the global max pooling process of PointNet. This information loss weakens the effectiveness of the distribution matching, preventing DM from accurately capturing the feature distribution of the original dataset. DC, on the other hand, aims to match the gradients of the network. However, PointNet is significantly larger than typical networks in image-based distillation tasks, making it difficult for DC to precisely align the gradients. As a result, it consistently underperforms compared to the proposed method. 
The proposed method outperforms both coreset selection and existing distillation techniques. When the PPC is small, accurately capturing the original distribution is challenging, leading less noticeable improvement. As the PPC increases, the proposed method more effectively matches the feature distributions significantly improving the performance.

\begin{table*}[t]
    \centering
    \begin{minipage}[t]{0.45\textwidth}
    \centering
    \setlength{\tabcolsep}{4.5pt}
    \caption{Comparison of cross-dataset generalization performance at the PPC value of 3, where the models are trained on a subset of ModelNet40 and evaluated on a subset of ShapeNet, and vice versa.}
    \label{tab:cross_dataset}
    \resizebox{\linewidth}{!}{%
        \begin{tabular}{c|ccccc}
        \toprule
        \textbf{Setting} & \textbf{DC} & \textbf{DM} & \textbf{PCC} & \textbf{Ours} \\
        \midrule
        \midrule
        \textbf{SN} $\rightarrow$ \textbf{MN40} & 54.43 & 49.66 & 57.43 & \textbf{61.54} \\
        \textbf{MN40} $\rightarrow$ \textbf{SN} & 53.75 & 47.35 & 49.42 & \textbf{54.40} \\
        \bottomrule
    \end{tabular}
    }
    \vspace{1.5em}
    \centering
    \addtocounter{table}{1}
    \caption{Comparison of training times (in hours) required for dataset distillation at the PPC value of 3.}
    \label{tab:tt}
    \resizebox{\linewidth}{!}{%
        \begin{tabular}{c|ccccc}
            \toprule
            \textbf{Method} & \textbf{MN10} & \textbf{MN40} & \textbf{SN} & \textbf{SONN} \\ 
            \midrule 
            \midrule
            \textbf{DC}            & 1.52  & 5.43 & 7.57 & 2.19 \\ 
            \textbf{DM}            & {\textbf {0.04}}  & {\textbf {0.11}} & {\textbf{0.15}} & {\textbf{0.05}}  \\
            \textbf{Ours} & 0.08  & 0.26 & 0.35 & 0.11 \\ 
            \bottomrule
            \end{tabular}
    }
    \end{minipage}\hfill
    \vspace{-1mm}
    \begin{minipage}[t]{0.53\textwidth}
        \centering
        \addtocounter{table}{-2}
        \caption{Performance comparison across different initialization strategies. The accuracies for PPC values of 1, 3, and 10 are averaged. \textbf{MN10}: ModelNet10. \textbf{MN40}: ModelNet40. 
        \textbf{SN}: ShapeNet. \textbf{SONN}: ScanObjectNN.}
        \renewcommand{\arraystretch}{0.95}
        \label{tab:init}
        \resizebox{\linewidth}{!}{%
        \begin{tabular}{@{}c|c|cccc@{}}
          \toprule
          \textbf{Init} & \textbf{Method} & \textbf{MN10} & \textbf{MN40} & \textbf{SN} & \textbf{SONN} \\ 
          \midrule
          \midrule
          
          \multirow{3}{*}{\textbf{Noise}} 
          & \textbf{DC}   & 18.77 &  6.05& 5.24 & 12.73  \\
          & \textbf{DM}   & 26.53 & 12.04&12.09&13.55\\
          & \textbf{Ours} & \textbf{70.69} & \textbf{65.25}& \textbf{58.44} & \textbf{29.03} \\
          \midrule
          \multirow{3}{*}{\textbf{Random}} 
          & \textbf{DC}   & 68.32 & 64.71 & 58.15 & 26.07\\ 
          & \textbf{DM}   & 65.17 & 56.50 & 49.39 & 24.57 \\ 
          & \textbf{Ours} & \textbf{72.48} & \textbf{69.31} & \textbf{60.76} & \textbf{31.01} \\
          \midrule
          \multirow{3}{*}{\textbf{K-center}}
          & \textbf{DC}   & 69.57 & 61.85 & 42.88 & 21.65 \\   
          & \textbf{DM}   & 63.04 & 49.93 & 44.82 & 21.59\\   
          & \textbf{Ours} & \textbf{72.72} & \textbf{68.20} & \textbf{60.70} & \textbf{28.25}\\   
          \midrule
          \multirow{3}{*}{\textbf{Herding}}
          & \textbf{DC\textsuperscript{†}} & 71.50 & 66.79 & 59.23 & 26.98  \\ 
          & \textbf{DM} & 64.76 & 59.16 & 52.61 & 27.58\\   
          & \textbf{Ours} & \textbf{72.93} & \textbf{69.67} & \textbf{62.25} & \textbf{31.08}\\  
          \bottomrule
        \end{tabular}
      } 
    
      \noindent\makebox[\textwidth][r]{%
        {† \fontsize{7}{7}\selectfont DC~\cite{gm} with Herding initialization corresponds to PCC~\cite{pcc}}%
      }
    \end{minipage}
\end{table*}

\subsubsection{Cross-Architecture Generalization}
We also present the cross-architecture generalization performance in Table~\ref{tab:tab2}, where the synthetic dataset is optimized using PointNet~\cite{pointnet} and evaluated on different backbone networks including PointNet++\cite{pointnet2}, DGCNN\cite{dgcnn}, PointConv~\cite{pointconv}, and PT~\cite{pointtransformer}, respectively. Unlike PointNet, the architectures used for generalization performance comparison follow hierarchical structure similar to CNN-based networks in image processing, exhibiting significantly different characteristics from those of PointNet. This difference makes it essential for the distilled dataset to retain rich and transferable features that generalize well toward different architectures.
Due to significant information loss caused by the pooling operation, DM primarily captures the features specific to the PointNet, failing to preserve the original feature distribution. Consequently, the synthetic dataset lacks diverse structural details, resulting in poor generalization performance, particularly on hierarchical models. DC, while performing better than DM, also suffers from key limitations. It matches the gradients obtained from a trained PointNet, the synthetic dataset becomes overfitted to that specific architecture and fails to generalize toward other networks.
Consequently, the generalization ability of DC degrades when applied to other architectures.
In contrary, the proposed method addresses these limitations by directly matching the feature distributions extracted from randomly initialized network. By comparing feature maps containing richer and less architecture-biased information, the synthetic dataset captures the features relevant to the PointNet while following the original data distribution more closely. Wherease DC suffers from the overfitting and DM loses critical information, the proposed method yields higher adaptability showing strong generalization performance across various backbone networks.


\subsubsection{Cross-Dataset Generalization}

To further evaluate the generalization capability of the distilled datasets across different domains, we provide an experimental result of cross-dataset experiment in Table \ref{tab:cross_dataset}. Specifically, we constructed the subsets of ShapeNet and ModelNet40, respectively, by selecting the 16 object classes shared between them. Then we performed the experiments in which the dataset was distilled on the subset of ShapeNet and tested on ModelNet40 (\textbf{SN} $\rightarrow$ \textbf{MN40}), and vice versa (\textbf{MN40} $\rightarrow$ \textbf{SN}). As shown in Table \ref{tab:cross_dataset}, the proposed method consistently achieves the best performance in both settings of the cross-dataset evaluation, indicating that our distilled datasets effectively capture geometric patterns that are useful regardless of the datasets.

\subsubsection{Initialization Strategies}
Table~\ref{tab:init} presents the classification performance of DC~\cite{gm}, DM~\cite{dm}, and PCC~\cite{pcc} under different initialization strategies, including uniform noise, random selection~\cite{herding(icarl)}, Herding~\cite{herding(e2e), herding(scail)}, and K-Center~\cite{kcenter}. Note that the results for PCC are obtained using our implementation, where PCC is equivalent to performing DC with Herding initialization.
When initialized with noise, DC fails to effectively optimize the synthetic dataset because the network has far more parameters than typical networks used for image dataset distillation, making it difficult to align the gradients. Therefore, the original dataset cannot properly guide the training process, leading to poor convergence. In contrast, the structured initializations such as random selection, Herding, or K-Center improve the performance, where the Herding performs the best by selecting the most representative samples. Unlike DC, DM directly matches feature representations rather than relying on gradients, making it less sensitive to initialization strategies. However, due to information loss, DM consistently underperforms DC, even when structured initializations are used.
In contrast, the proposed method achieves favorable performance even when initialized from uniform noise. Furthermore, it significantly outperforms other approaches under heuristic initialization strategies, such as random selection. These results demonstrate that the proposed method is robust across diverse initialization schemes and effectively addresses the instability issues inherent in existing dataset distillation methods.

\subsubsection{Training Time}
Table~\ref{tab:tt} compares the training times between the proposed method and the existing methods. DM~\cite{dm} achieves the fastest training time as it uses only a small subset of the overall features, however it often ignores many useful features leading to performance degradation. In contrast, DC~\cite{gm} incurs a significantly higher computational cost as it aligns the gradients across all the parameters. Notably, for the largest dataset of ShapeNet~\cite{shapenet}, DC requires over 7 hours for training 50 times slower that DM that completes the training in just 9 minutes (0.15 hours). Since the proposed method is based on the distribution matching approach, its training time is marginally increased compared to DM, yet achieves significant improvement in dataset distillation performance. To further justify the computational overhead, we conducted an experiment where DM was trained with its default setting, whereas the proposed method was trained during the same training time as DM. As shown in Table~\ref{tab:comp_over}, even under the same training time condition, our method consistently outperforms DM.

\begin{table*}[t]
	\centering
	\begin{minipage}[t]{0.48\textwidth} 
		\centering
		\setlength{\tabcolsep}{5.5pt}
		\addtocounter{table}{1}
		\caption{Performance comparison between DM and the proposed method at the PPC value of 3, where the proposed method was trained during the same time as DM.}
		\label{tab:comp_over}
		\begin{tabular}{c|cccc}
			\toprule
			\textbf{Method} & \textbf{MN10} & \textbf{MN40} & \textbf{SN} & \textbf{SONN} \\ \midrule \midrule
			\textbf{DM}       & 77.58 & 62.04 & 56.14 & 23.33 \\
			\textbf{Ours}     & \textbf{83.64} & \textbf{70.56} & \textbf{63.34} & \textbf{33.06} \\
			\bottomrule
		\end{tabular}
		\vspace{-2mm}
	\end{minipage}
	\hfill
	\begin{minipage}[t]{0.5\textwidth}
		\centering
		\caption{Ablation study of the proposed semantic alignment~({\bf SA}) method. Experiments were performed at the PPC value of 3.}
		\label{tab:sadm}
		\begin{tabular}{c|cccc}
			\toprule
			\textbf{Method} & \textbf{MN10} & \textbf{MN40} & \textbf{SN} & \textbf{SONN} \\ 
			\midrule
			\midrule
			Random        & 75.17 & 59.96 & 54.84 & 20.42 \\
			w/o {\bf SA}  & 75.01 & 59.96 & 53.91 & 20.20 \\
			w/ {\bf SA}   & \textbf{84.42} & \textbf{72.08} & \textbf{63.74} & \textbf{31.84} \\
			\bottomrule 
		\end{tabular}
		\vspace{-2mm}
	\end{minipage}
\end{table*}

\subsubsection{Qualitative Comparison of Synthetic Datasets}
\begin{figure*}[t]
	\centering
	\includegraphics[width=1\linewidth]{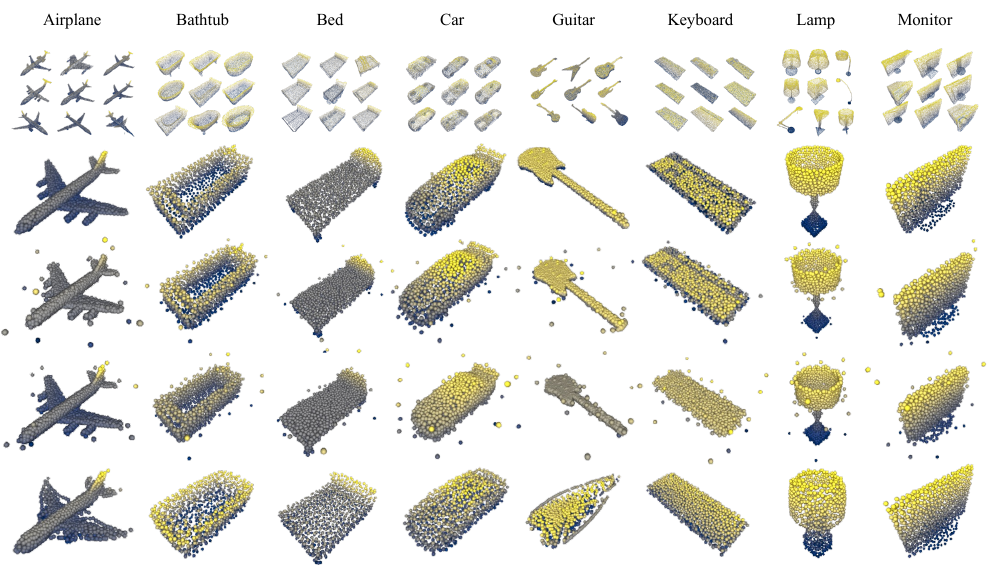}
	\caption{Comparison of synthetic datasets distilled from the ModelNet40~\cite{modelnet40} dataset. Point clouds were colorized according to the y-coordinates. From top to bottom, the orginal dataset (first), initialized point cloud models (second), and the distilled synthetic datasets by using DC~\cite{gm} (third), DM~\cite{dm} (fourth), and the proposed method (fifth).}
	\label{fig:fig3} \vspace{-2mm}
\end{figure*}

Figure~\ref{fig:fig3} compares the resulting synthetic datasets distilled by using the proposed method and the existing methods, respectively. The first row shows the original datasets and the second row shows initialized point cloud objects. DC~\cite{gm} fails to deviate significantly from the initialized objects and causes noise. Similarly, DM~\cite{dm} also maintains the original shape of the initialized objects while shifting certain points only, that hinders to capture meaningful structural changes. In contrast, as shown in the bottom row, the proposed method successfully preserves the overall semantic structures of 3D objects in each class, while selectively learning essential features. For example, in the airplane class, we observe significant changes in the edges and corners, that are critical for class discrimination, especially around the wings. Also, in the guitar class, the proposed method jointly optimizes the shape and orientation of the synthesized 3D objects. 

\subsection{Ablation Study}

\subsubsection{Effect of SADM Loss}
Table~\ref{tab:sadm} presents an ablation study on the impact of the proposed SADM loss on classification performance using PointNet~\cite{pointnet}. Without preserving semantic alignment, the unordered nature of point clouds restricts effective feature matching, resulting in performance nearly identical to that of random selection~\cite{herding(icarl)}. This indicates that the model has not been properly optimized. In contrast, when semantic alignment is applied, accuracy improves across all datasets, as the aligned features provide a consistent and structured representation that facilitates more effective feature matching during optimization. These results confirm that preserving inherent semantic alignment via feature sorting enables the network to exploit well-aligned corresponding features, significantly enhancing classification performance.

\subsubsection{Effect of Optimal Rotation Estimation}

Table~\ref{tab:ore} presents the ablation study to evaluate the impact of joint optimization of synthetic dataset and rotation angles. We categorized the test datsets into the aligned, mixed, and rotated groups to analyze the effectiveness of the proposed rotation estimation. The aligned group consists of the datasets where the objects maintain consistent orientations, including ModelNet10 and ShapeNet~\cite{shapenet}. The mixed group includes ModelNet40~\cite{modelnet40} where only certain classes exhibit rotation variations. The rotated group is  ScanObjectNN~\cite{scanobjectnn} where the objects exhibit arbitrary rotations across all classes. 
The baseline (first row) shows the performance without optimization which is identical to that of the random selection.
In the second row, we present the results where only the orientation of the synthetic data is optimized using the proposed SADM loss, without updating the synthetic dataset. This leads to performance improvements across all the groups compared to random selection. The improvement becomes more significant as the classes with rotation variation become more dominant. Optimizing the synthetic dataset further boosts the performance, with the best results achieved when both the synthetic data and rotation angles are optimized simultaneously. In particular, the accuracy on the rotated dataset improves significantly, demonstrating that the proposed joint optimization enables the model to be resilient to rotation variations.

\subsubsection{Effect of Point Sorting}
We compared the performance of the proposed SADM against two representative point sorting schemes. The axis-aligned sorting method simply orders the points in the ascending order of their $z$-coordinates, without considering any structural or semantic relationships. The Z-order sorting method~\cite{morton} assigns an index to each point based on its 3D coordinates simultaneously, ensuring that points close in space receive nearby indices. As shown in Table~\ref{tab:pss}, the existing sorting methods yield only marginal improvements over the baseline that does not use point sorting. Moreover, they fail to provide consistent semantic alignment across different samples. In contrast, the proposed SADM significantly outperforms these methods by achieving semantically consistent matching, thereby enhancing the effectiveness of dataset distillation for 3D point clouds.

\begin{table*}[t]
	\centering
	\begin{minipage}[t]{0.43\textwidth} 
		\centering
		\caption{Ablation study of the proposed optimal rotation estimation. 
			Experiments were performed at the PPC value of 3.}
		\label{tab:ore}
		\begin{tabular}{cc|ccc}
			\toprule
			${\boldsymbol{\mathcal S}}$ & ${\boldsymbol{\mathcal {\theta}}}$ 
			& \textbf{Aligned} & \textbf{Mixed} & \textbf{Rotated} \\
			\midrule
			\midrule
			\_         & \_          & 65.01 & 59.96 & 20.42 \\
			\_         & \checkmark  & 66.22 & 62.17 & 24.89 \\
			\checkmark & \_          & 73.96 & 71.51 & 29.72 \\
			\checkmark & \checkmark  & \textbf{74.35} & \textbf{72.08} & \textbf{31.84} \\
			\bottomrule
		\end{tabular}
		\vspace{-2mm}
	\end{minipage}
	\hfill
	\begin{minipage}[t]{0.55\textwidth}
		\centering
		\setlength{\tabcolsep}{6pt}
		\caption{Performance comparison of different point sorting schemes at the PPC value of 3. 
			For fair comparison, the rotation parameter optimization was not applied.}
		\label{tab:pss}
		\begin{tabular}{c|cccc}
			\toprule
			\textbf{Method} & \textbf{MN10} & \textbf{MN40} & \textbf{SN} & \textbf{SONN} \\
			\midrule
			\midrule
			Unsorted      & 75.21 & 59.96 & 53.91 & 20.20 \\
			Axis-Aligned  & 76.44 & 60.85 & 53.97 & 21.80 \\
			Z-order~\cite{morton} & 76.55 & 61.47 & 55.17 & 21.76 \\
			SADM          & \textbf{83.17} & \textbf{67.52} & \textbf{60.71} & \textbf{27.66} \\
			\bottomrule
		\end{tabular}
		\vspace{-2mm}
	\end{minipage}
\end{table*}

\subsection{Limitation} 
While the proposed SADM loss enhances the semantic consistency by sorting the features, it does not completely align semantically meaningful regions across different point cloud objects. Estimating optimal rotations also increases the complexity, particularly when the datasets are already well-aligned to the canonical axes. 

\section{Conclusion}
We proposed a semantically aligned and orientation-aware dataset distillation framework for 3D point clouds. To address the inconsistency in point ordering between compared 3D objects, we devised a Semantically Aligned Distribution Matching (SADM) loss that compares sorted features within each channel. Additionally, we introduced learnable rotation angle parameters to estimate the optimal orientation of synthetic objects. The geometric structures and orientations of the synthetic objects are jointly optimized during dataset distillation. Experimental results on four widely used benchmark datasets—ModelNet10~\cite{modelnet40}, ModelNet40~\cite{modelnet40}, ShapeNet~\cite{shapenet}, and ScanObjectNN~\cite{scanobjectnn}—demonstrate that the proposed method outperforms existing distillation approaches while maintaining strong cross-architecture generalization. Furthermore, the effectiveness of each component is validated through extensive ablation studies.

\section*{Acknowledgement}
This work was supported in part by the National Research Foundation of Korea (NRF) Grant funded by Korea Government [Ministry of Science and ICT (MSIT)] under Grant RS-2024-00392536; in part by the Institute of Information and Communications Technology Planning and Evaluation (IITP) Grant funded by Korea Government (MSIT) (Leading Generative AI Human Resources Development) under Grant IITP-2025-RS-2024-00360227; in part by the Artificial Intelligence Gradate School Program, Ulsan National Institute of Science of Technology, under Grant RS-2020-II201336; and in part by the Artificial Intelligence Innovation Hub under Grant RS-2021-II212068.

{
    \small
    \bibliographystyle{plainnat}
    \bibliography{main}
}

\newpage

\appendix

\section*{\LARGE Appendix}

\section{Proof of Proposition 1}

\textbf{Proposition 1.}
\textit{Jointly optimizing the synthetic dataset $\boldsymbol{\mathcal{S}}$ and the rotation parameters $\boldsymbol{\theta} = (\theta_x, \theta_y, \theta_z)$ guarantees a lower or equal loss to that of optimizing $\boldsymbol{\mathcal{S}}$ alone.}
\begin{equation}
\min_{\left\{\boldsymbol{\mathcal{S}}, \boldsymbol{\theta}\right\}} \mathcal{L}_{\text{SADM}}(\boldsymbol{\mathcal{T}}, \mathcal{R}_{\boldsymbol{\theta}}(\boldsymbol{\mathcal{S}})) \leq \min_{\boldsymbol{\mathcal{S}}} \mathcal{L}_{\text{SADM}}(\boldsymbol{\mathcal{T}}, \boldsymbol{\mathcal{S}}),
\end{equation}
\textit{where $\mathcal{R}_{\boldsymbol{\theta}}$ denotes the rotation operator according to $\boldsymbol{\theta}$.}

\textbf{\textit{Proof. }} Let us consider two optimization objectives. The first is the baseline optimization, which optimizes only the synthetic dataset without applying any rotation
\begin{equation}
	L := \min_{\boldsymbol{\mathcal{S}}} \mathcal{L}_{\text{SADM}}\left(\boldsymbol{\mathcal{T}},\, \boldsymbol{\mathcal{S}}\right).
\end{equation}
The second is the proposed joint optimization over both the synthetic dataset $\boldsymbol{\mathcal{S}}$ and the rotation parameters $\boldsymbol{\theta} = (\theta_x, \theta_y, \theta_z)$, given
\begin{equation}
L^\star := \min_{\left\{\boldsymbol{\mathcal{S}},\, \boldsymbol{\theta}\right\}} \mathcal{L}_{\text{SADM}}\left(\boldsymbol{\mathcal{T}},\, \mathcal{R}_{\boldsymbol{\theta}}(\boldsymbol{\mathcal{S}})\right).
\end{equation}

Consider a special case of the proposed joint optimization framework where no rotation is applied, i.e., the rotation parameters are fixed to $\boldsymbol{\theta} = (0,0,0)$. In such a case, $\mathcal{R}_{\boldsymbol{\theta}}(\boldsymbol{\mathcal{S}}) = \boldsymbol{\mathcal{S}}$, and the baseline optimization can be regarded as a special case of the proposed joint optimization. This implies that the feasible set of the baseline optimization is a subset of that of the joint optimization such that
\begin{equation}
\left\{ (\boldsymbol{\mathcal{S}},\, \boldsymbol{\theta}) \,\middle|\, \boldsymbol{\theta} = (0,0,0) \right\} \subseteq \left\{ (\boldsymbol{\mathcal{S}},\, \boldsymbol{\theta}) \right\}.
\end{equation}

The joint optimization is performed over a larger feasible set than that of the baseline optimization, and therefore its optimal solution must not be higher than that of the baseline. Therefore,
\begin{equation}
L^\star \leq L.
\end{equation}


\section{Expeimental Details}

{\bf Datasets. } Table \ref{tab:supp_tab1} summarizes the number of training and testing samples for ModelNet10~\cite{modelnet40}, ModelNet40~\cite{modelnet40}, ShapeNet~\cite{shapenet}, and ScanObjectNN~\cite{scanobjectnn}.

\begin{itemize}
    \item \textbf{ModelNet10} consists of 10 categories of aligned 3D CAD models. There is no rotational variation.
    \item \textbf{ModelNet40} includes 40 categories from the same source to ModelNet10, with partially misaligned instances introducing moderate rotational variation.
    \item \textbf{ShapeNet} contains 55 categories of manually aligned 3D models without rotational variation.
    \item \textbf{ScanObjectNN} consists of 15 categories from real-world RGB-D scans with background clutter, occlusion, and sensor noise. Objects are unaligned, and rotation variation is inherent. We use the \textbf{PB\_T50\_RS} split, which includes perturbed background, translation jitter, rotation, and scaling.
\end{itemize}

\begin{table}[hb]
	\centering
	\caption{Train/test statistics for ModelNet10, ModelNet40, ShapeNet and ScanObjectNN.}
	\begin{tabular}{ccccc}
		\toprule
		&ModelNet10& ModelNet40 & ShapeNet & ScanObjectNN\\
		\midrule
		\# of training samples   & 3991& 9843& 35708  & 11416 \\
		\# of test samples  & 908 & 2468& 10261  & 2882 \\
		\# of classes &10&40&55&15\\
		\# of points for each sample &1024&1024&1024&1024\\
		\bottomrule 
	\end{tabular}
	\label{tab:supp_tab1}
\end{table}

{\bf Network Architecture. }We evaluate our method on five representative point-based architectures: PointNet~\cite{pointnet}, PointNet++~\cite{pointnet2}, DGCNN~\cite{dgcnn}, PointConv~\cite{pointconv}, and Point Transformer~\cite{pointtransformer}.

\begin{itemize}

    \item \textbf{PointNet} is a widely used neural network designed for processing 3D point clouds. We use PointNet as the backbone in dataset distillation model. While the original PointNet includes two transformation modules for aligning input and features, we only use the input transformation module. The rest of the architecture follows the standard PointNet design.

    \item \textbf{PointNet++} is used for cross-architecture evaluation. To reduce computational cost and accelerate training, we decrease the width of all MLP layers by half, while keeping the overall structure unchanged. We use two set abstraction modules with multi-scale grouping and one with single-scale grouping to extract point features, followed by two linear layers and a final classifier for prediction.

    \item \textbf{DGCNN} employs dynamic graph construction and EdgeConv layers to capture local geometric relationships by recomputing a $k$-NN graph ($k=20$) at each layer in the feature space. We use the default architecture, which consists of four EdgeConv layers followed by two linear layers and a final classifier.

    \item \textbf{PointConv} extends convolution to point clouds by accounting for non-uniform point densities during feature learning. The architecture consists of three density-aware set abstraction layers, followed by two linear layers and a final classifier.

    \item \textbf{Point Transformer} applies self-attention mechanisms to point clouds to capture both local and global dependencies. We use the default configuration with four transformer blocks, followed by two linear layers and a final classifier.
\end{itemize}

{\bf Implementation Details. }We optimized the synthetic dataset $\boldsymbol{\mathcal{S}}$ using stochastic gradient descent (SGD) with a learning rate of 0.01, a momentum of 0.5, a weight decay of 0, and a batch size of 8 per class sampled from the original dataset $\boldsymbol{\mathcal{T}}$, while the batch size of the synthetic dataset was set equal to the number of synthetic samples per class (PPC). The synthetic dataset optimization was performed for 1,500 iterations, and the corresponding configuration is summarized in Table~\ref{tab:supp_tab2-a}. To preserve fine-grained geometric details, the SADM loss was computed using the feature maps before the max pooling layer. Additionally, a secondary loss term was applied to the top-1 sorted feature values in each channel to emphasize semantically dominant regions. The loss weights $\lambda_1$ and $\lambda_2$ were determined based on PPC, and set to 0.002, 0.006, and 0.02 for $\lambda_1$, and 0.001, 0.003, and 0.01 for $\lambda_2$ when PPC was 1, 3, and 10, respectively.

The rotation parameters $\boldsymbol{\theta} = (\theta_x, \theta_y, \theta_z)$ were jointly optimized with the synthetic dataset using SGD with a momentum of 0.5 and a weight decay of 0. The learning rates were set to 0.5 for $\theta_x$ and $\theta_z$, and 5.0 for $\theta_y$, reflecting that the samples in datasets are vertically aligned. A step decay scheduler was used with a step size of 100 and a decay factor of 0.5. This setup is detailed in Table~\ref{tab:supp_tab2-b}.

For evaluation, we trained all backbone networks including PointNet, PointNet++, DGCNN, PointConv, and Point Transformer using SGD with a learning rate of 0.01, a momentum of 0.9, and a weight decay of 0.0005. A step decay scheduler was applied with a step size of 250 and a decay factor of 0.1. Each model was trained for 500 epochs with a batch size of 8. These test-time settings are listed in Table~\ref{tab:supp_tab2-c}.

All experiments were conducted on a single NVIDIA GeForce RTX 3090 GPU.

\begin{table}[ht]
  \centering
  \caption{Hyperparameter settings used for (a) optimizing the synthetic dataset, (b) optimizing the rotation parameters, and (c) evaluation network.}
  \vspace{1em}
  \scriptsize
  \begin{minipage}[t]{0.39\columnwidth}
    \centering
    \begin{tabular}{@{}c|ccc@{}}
      \toprule
      Hyperparameters & PPC1 & PPC3 & PPC10 \\ 
      \midrule
      Optimizer     & SGD  & SGD  & SGD  \\ 
      Momentum      & 0.5  & 0.5  & 0.5  \\
      Weight Decay  & 0.0  & 0.0  & 0.0  \\
      Learning Rate & 0.01 & 0.01 & 0.01 \\
      Iteration     & 1500 & 1500 & 1500 \\
      \midrule
      $\lambda_1$ & 0.002 & 0.006 & 0.02 \\ 
      $\lambda_2$ & 0.001 & 0.003 & 0.01 \\
      \midrule       
      Batch Size~$(\boldsymbol{\mathcal{T}})$ & 8 & 8 & 8   \\
      \bottomrule
    \end{tabular}
    \subcaption{}
    \label{tab:supp_tab2-a}
  \end{minipage}
   \hspace{0.0\columnwidth}%
  \begin{minipage}[t]{0.3\columnwidth}
    \centering
    \begin{tabular}{@{}c|c@{}}
      \toprule
      \multicolumn{2}{c}{Hyperparameters} \\
      
      \midrule
      
      Optimizer    & SGD    \\
      Momentum     & 0.5    \\
      Weight Decay & 0.0    \\
      
      Learning Rate~$(\theta_x)$ & 0.5    \\
      Learning Rate~$(\theta_y)$ & 5.0    \\
      Learning Rate~$(\theta_z)$ & 0.5    \\
      
      \midrule
      
      Scheduler & StepLR \\
      Step Size & 100    \\
      Gamma     & 0.5    \\
      
      \bottomrule
    \end{tabular}
    \subcaption{}
    \label{tab:supp_tab2-b}
  \end{minipage}
   \hspace{0.0\columnwidth}%
  \begin{minipage}[t]{0.3\columnwidth}
    \centering
    \begin{tabular}{@{}c|c@{}}
      \toprule
      \multicolumn{2}{c}{Hyperparameters} \\
      \midrule
      Optimizer     & SGD    \\
      Momentum      & 0.9    \\
      Weight Decay  & 0.0005    \\
      Learning Rate & 0.01   \\
      Epochs        & 500   \\
      Batch Size    & 8      \\
      \midrule
      Scheduler    & StepLR \\
      Step Size    & 250    \\
      Gamma        & 0.1    \\
      \bottomrule 
    \end{tabular}
    \subcaption{}
    \label{tab:supp_tab2-c}
  \end{minipage}

  \vspace{1 ex}
  \label{tab:supp_tab2}
\end{table}

\begin{figure*}[t]
	\centering
        \includegraphics[width=\linewidth]{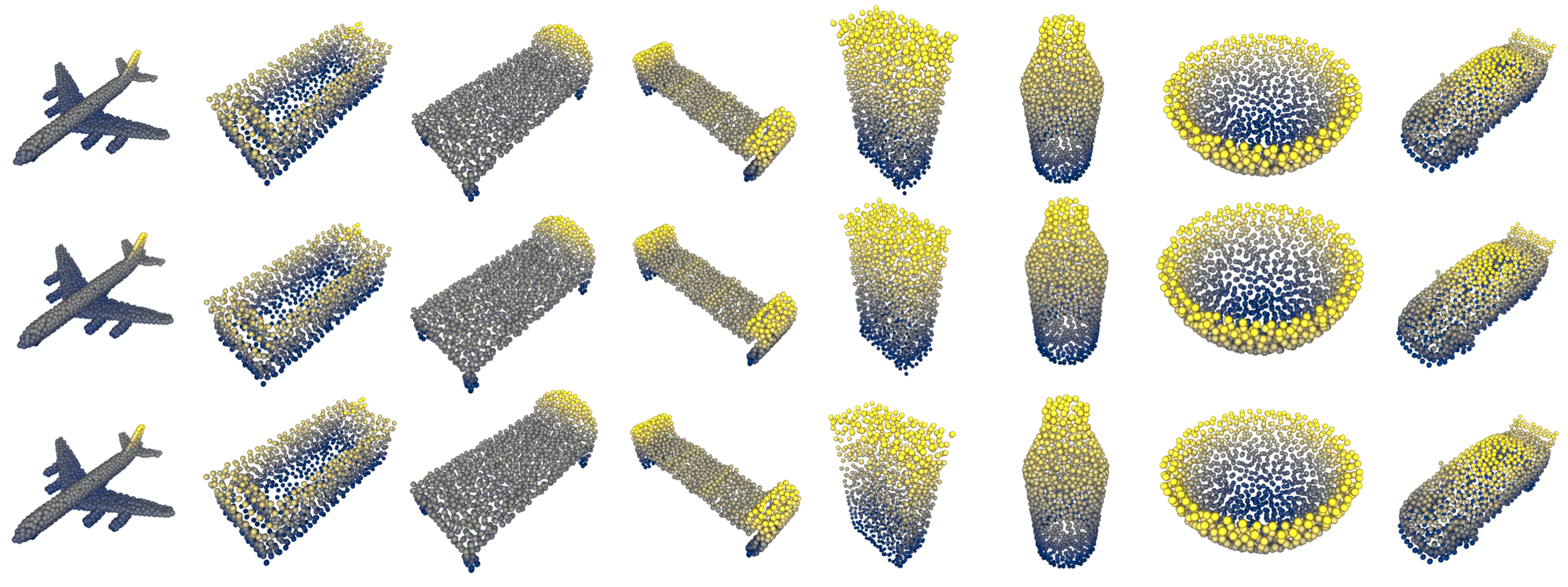}
        \caption{Visualization of synthetic samples of ModelNet40~\cite{modelnet40} at PPC 1, obtained by using the random selection~\cite{herding(icarl)}(top), DataDAM~\cite{datadam}(middle), and MTT~\cite{mtt1}(bottom). }
        \label{fig:supp_fig1}
\end{figure*}

{\bf Baselines. } 
All baseline methods were implemented on 3D point clouds using PointNet as the feature extractor, following their original designs. All methods were trained for 1,500 iterations using a batch size of 8 per class from the original dataset~$\boldsymbol{\mathcal{T}}$, and the training configurations are summarized in Table~\ref{tab:supp_tab3}. Details for each method are as follows

\begin{itemize}
    \item \textbf{DC}~\cite{gm}: We followed the original framework and performed gradient matching. The number of inner and outer loop steps were set to 1/1/10 and 1/5/3 for PPC values of 1, 3, and 10, respectively

    \item \textbf{DM}~\cite{dm}: Following the original setup, we used the feature vector obtained before the final classifier to compute the matching loss between the original and synthetic datasets.

    \item \textbf{MTT}~\cite{mtt1}: We followed the original framework for trajectory matching, with modifications only in the hyperparameter settings.
    
    \item \textbf{DataDAM}~\cite{datadam}: We followed the original framework for feature matching, making only minor adjustments to hyperparameters.
    
\end{itemize}

\begin{table}[ht]
    \centering
    \caption{Hyperparameter settings of the baselines.}
    \begin{tabular}{ccccc}
    \toprule
    & DC & DM & MTT & DataDAM \\ \midrule
    Mode & PointNet & PointNet & PointNet & PointNet \\
    Learning Rate & 0.0001 & 1 & 0.0001 & 0.0001 \\
    Batch Size~$(\boldsymbol{\mathcal{T}})$ & 8 & 8 & 8 & 8\\
    Iteration & 1500 & 1500 & 1500 & 1500  \\ 
    Inner Loop & 1 / 1 / 10 & -- & -- & -- \\
    Outer Loop & 1 / 5 / 3  & -- & -- & -- \\
    \bottomrule 
    \end{tabular}
    \label{tab:supp_tab3}
\end{table}


\section{Comparison with Additional Baselines}

Table~\ref{tab:supp_tab4} presents the additional comparison results of the classification accuracy with DataDAM~\cite{datadam} and MTT~\cite{mtt1}. For DataDAM and MTT, we used a possible learning rate to prevent divergence on the synthetic datasets. The performance of both DataDAM and MTT is almost similar to that of random selection~\cite{herding(icarl)}, as shown in the second and third rows in Figure~\ref{fig:supp_fig1} where most points remain nearly unchanged from their random initialization in the first row.

\begin{table}[ht]
    \centering
    \caption{Classification accuracy of the proposed method compared with DataDAM~\cite{datadam} and MTT~\cite{mtt1} initialized with randomly selected samples. }
    \resizebox{\textwidth}{!}{%
    \begin{tabular}{@{}c|ccc|ccc|ccc|ccc@{}}
    \toprule
    \textbf{Datasets} & \multicolumn{3}{c|}{\textbf{ModelNet10}} & \multicolumn{3}{c|}{\textbf{ModelNet40}} & \multicolumn{3}{c|}{\textbf{ShapeNet}} & \multicolumn{3}{c}{\textbf{ScanObjectNN}} \\ 
    \midrule
    \textbf{PPC} & 1 & 3 & 10 & 1 & 3 & 10 & 1 & 3 & 10 & 1 & 3 & 10 \\
    \midrule
    \textbf{MTT} & 28.95& 76.73& 85.19& 34.42& 59.81& 73.88& 33.90& 52.92& 62.73& 13.94& 20.28& 34.07 \\ 
    \textbf{DataDAM} & 40.51& 76.60& 86.56& 34.54& 60.37& 74.94& 35.70& 54.56& 63.64 & 16.29& 20.51& 35.45\\
    \midrule    
    \textbf{Ours} &\textbf{44.70} & \textbf{84.96} &\textbf{87.79} &\textbf{55.80}& \textbf{72.08}& \textbf{80.07}& \textbf{50.20}& \textbf{63.74}& \textbf{68.35}& \textbf{17.29}& \textbf{31.84}& \textbf{43.91} \\
    \bottomrule
    \end{tabular}
     }
     \label{tab:supp_tab4}
\end{table}

\section{Additional Ablation Studies}

To further support the effectiveness of the proposed rotation optimization method, we compared widely used rotation augmentation that applies random rotations to the training data to address the rotation variation of 3D point clouds. While this strategy is effective when dealing with datasets showing severe rotational variations, such as ScanObjectNN, it does not achieve good performance on datasets where the 3D objects are fully or partially aligned, such as ModelNet10, ShapeNet, and ModelNet40.
As shown in Table~\ref{tab:supp_tab5}, the proposed learnable rotation estimation method consistently improves the performance across all datasets, however, the random rotation augmentation introduces unnecessary variation in already aligned datasets, degrading the performance. This highlights the benefit of explicitly modeling the orientation via optimization over the naive augmentation.

\begin{table}[ht]
    \centering
    \caption{Performance comparison between the proposed rotation optimization and the random rotation augmentation (Aug.). All experiments were conducted at PPC 3.}
    \begin{tabular}{cc|ccc}
    \toprule
    \textbf{Prop.} & \textbf{Aug.} & \textbf{Aligned} & \textbf{Mixed} & \textbf{Rotated}\\
    \midrule
    \midrule
    \_         & \_         & 65.01 & 59.96 & 20.42\\
    \_         & \checkmark & 58.92 & 52.56 & 31.42\\
    \checkmark & \_         & \textbf{74.35} &\textbf{72.08} & \textbf{31.84} \\
    \bottomrule 
    \end{tabular}
    \label{tab:supp_tab5}
\end{table}

\section{Additional Qualitative Results}

\begin{figure*}[h]
	\centering
        \includegraphics[width=\textwidth]{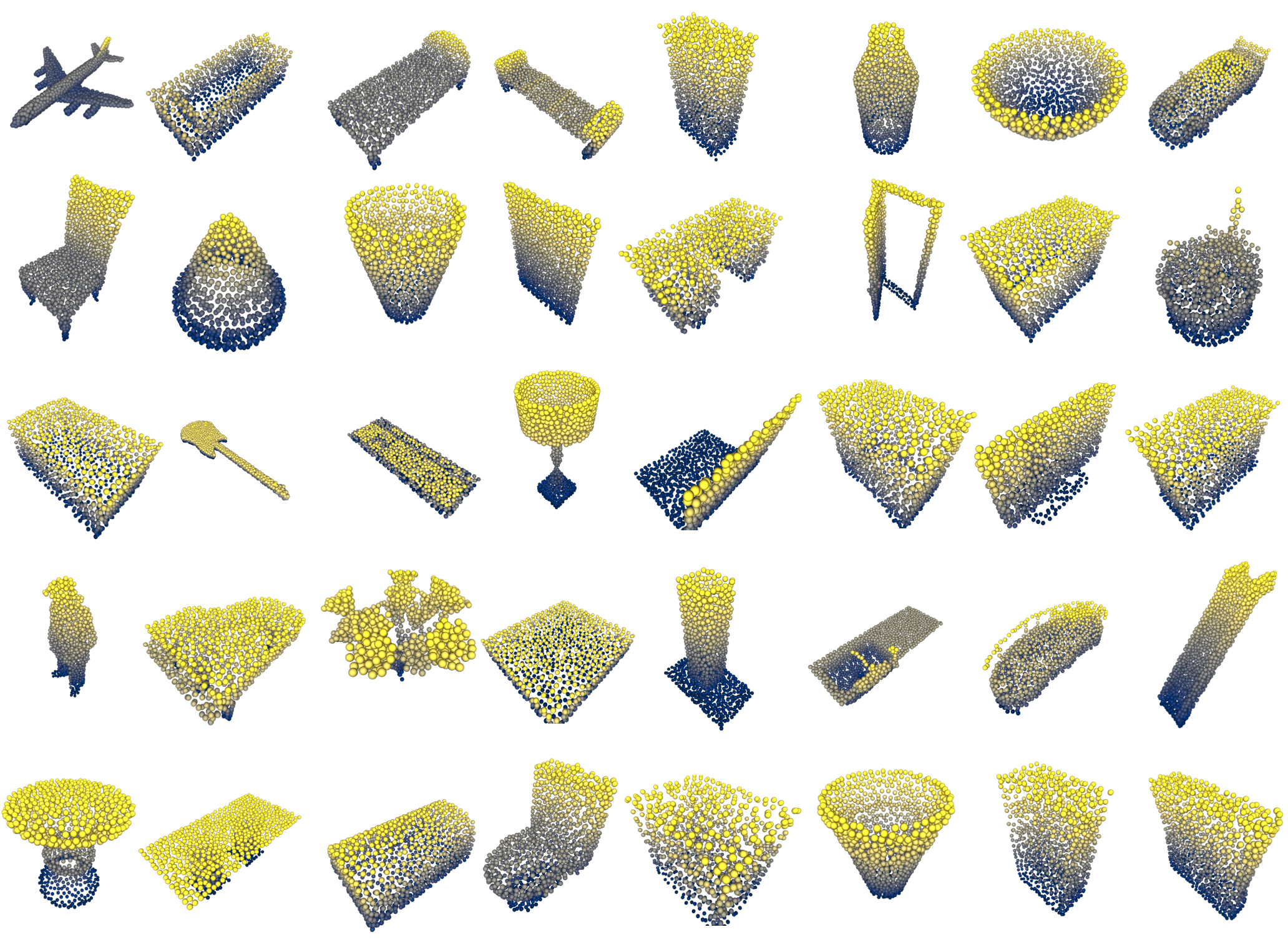}
        \caption{Randomly sampled 3D models of all classes from ModelNet40~\cite{modelnet40}.}
        \label{fig:supp_fig2}
\end{figure*}

\begin{figure*}[t]
	\centering
        \includegraphics[width=\textwidth]{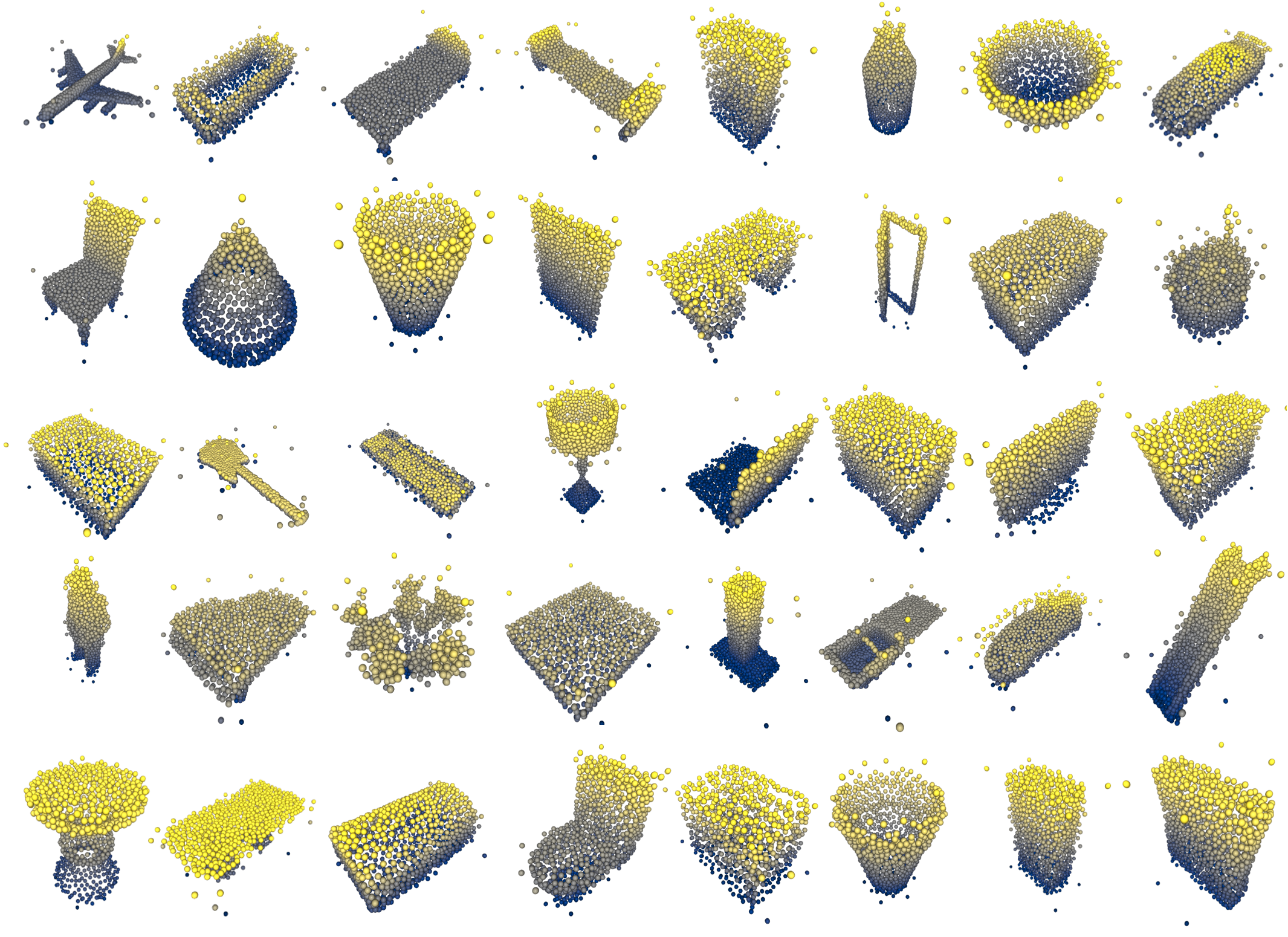}
        \caption{The synthetic dataset distilled from ModelNet40 by using DC at PPC 1.}
        \label{fig:supp_fig3}
\end{figure*}

\begin{figure*}[t]
	\centering
        \includegraphics[width=\textwidth]{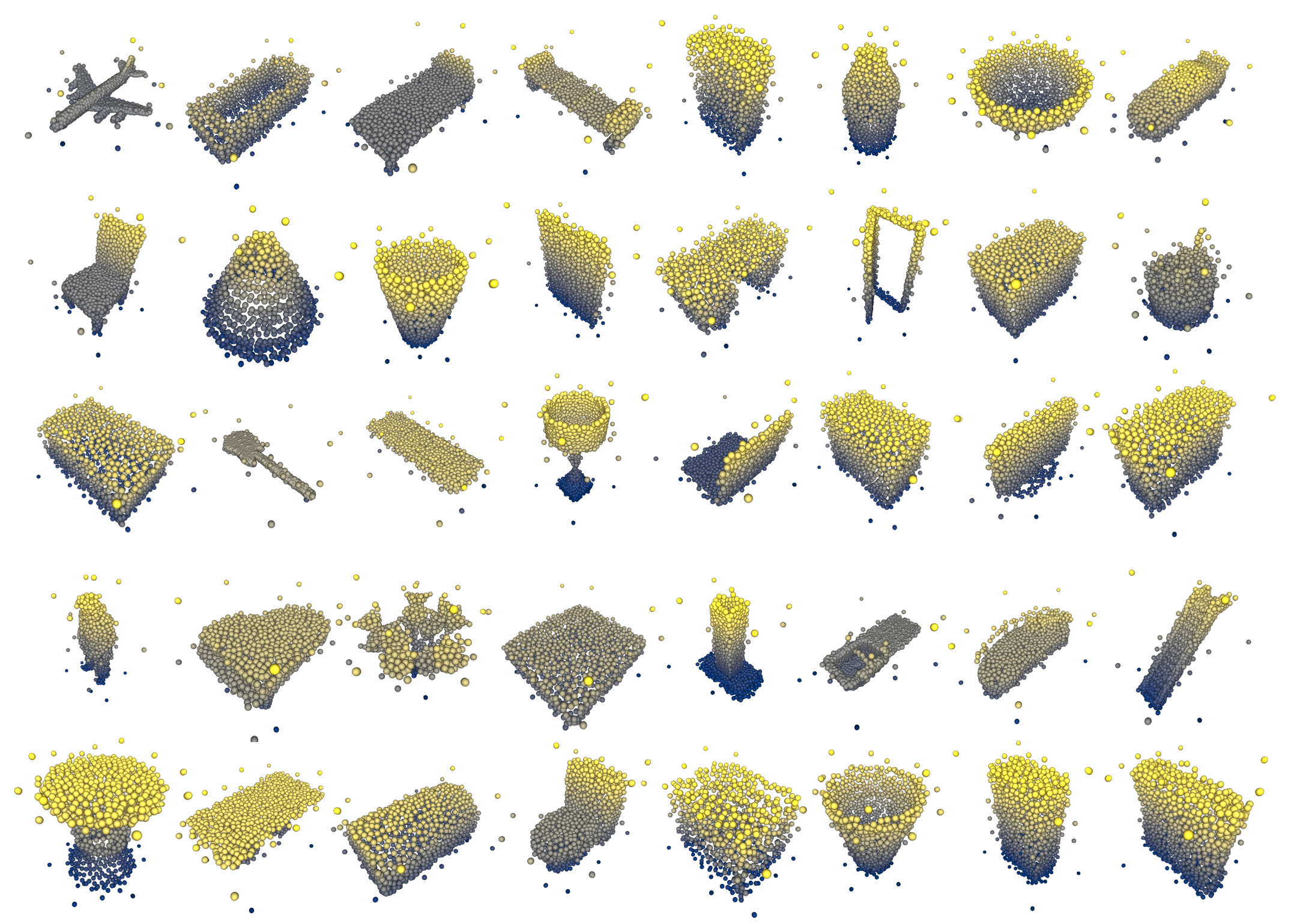}
        \caption{The synthetic dataset distilled from ModelNet40 by using DM at PPC 1.}
        \label{fig:supp_fig4}
\end{figure*}

\begin{figure*}[t]
	\centering
        \includegraphics[width=\textwidth]{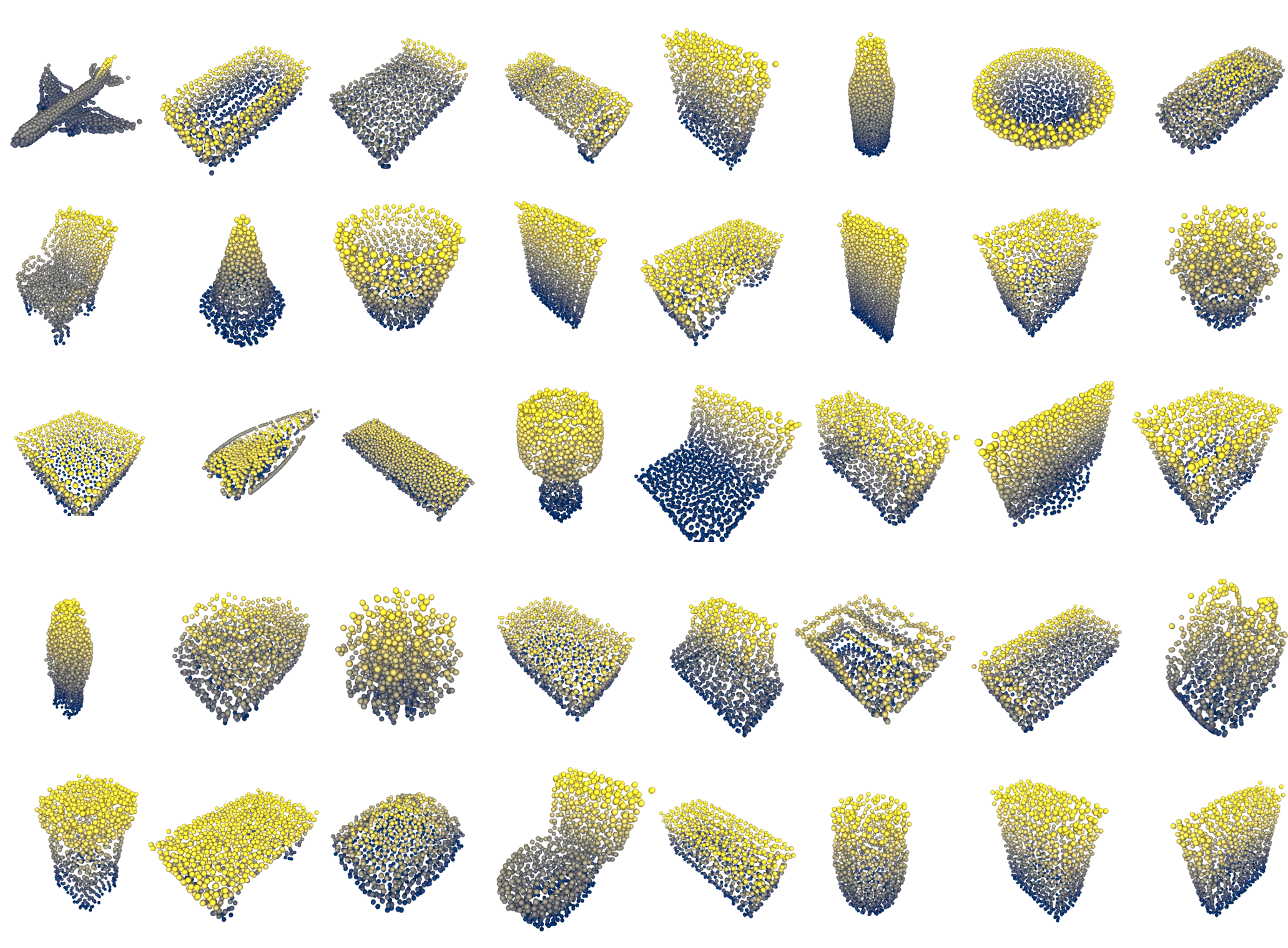}
        \caption{The synthetic dataset distilled from ModelNet40 by using the proposed method at PPC 1.}
        \label{fig:supp_fig5}
\end{figure*}

Figure~\ref{fig:supp_fig2} shows randomly sampled 3D models from ModelNet40. Figures~\ref{fig:supp_fig3}, \ref{fig:supp_fig4}, and \ref{fig:supp_fig5} visualize the synthesized models at PPC 1 obtained by using DC, DM, and the proposed method, respectively. While DC and DM maintain the geometric shapes of the original data with only slight movement of some points, the proposed method generates noise-free samples with high visual quality across all classes.

\begin{figure*}[t]
        \centering
        \subfloat[]{\includegraphics[width=\textwidth]{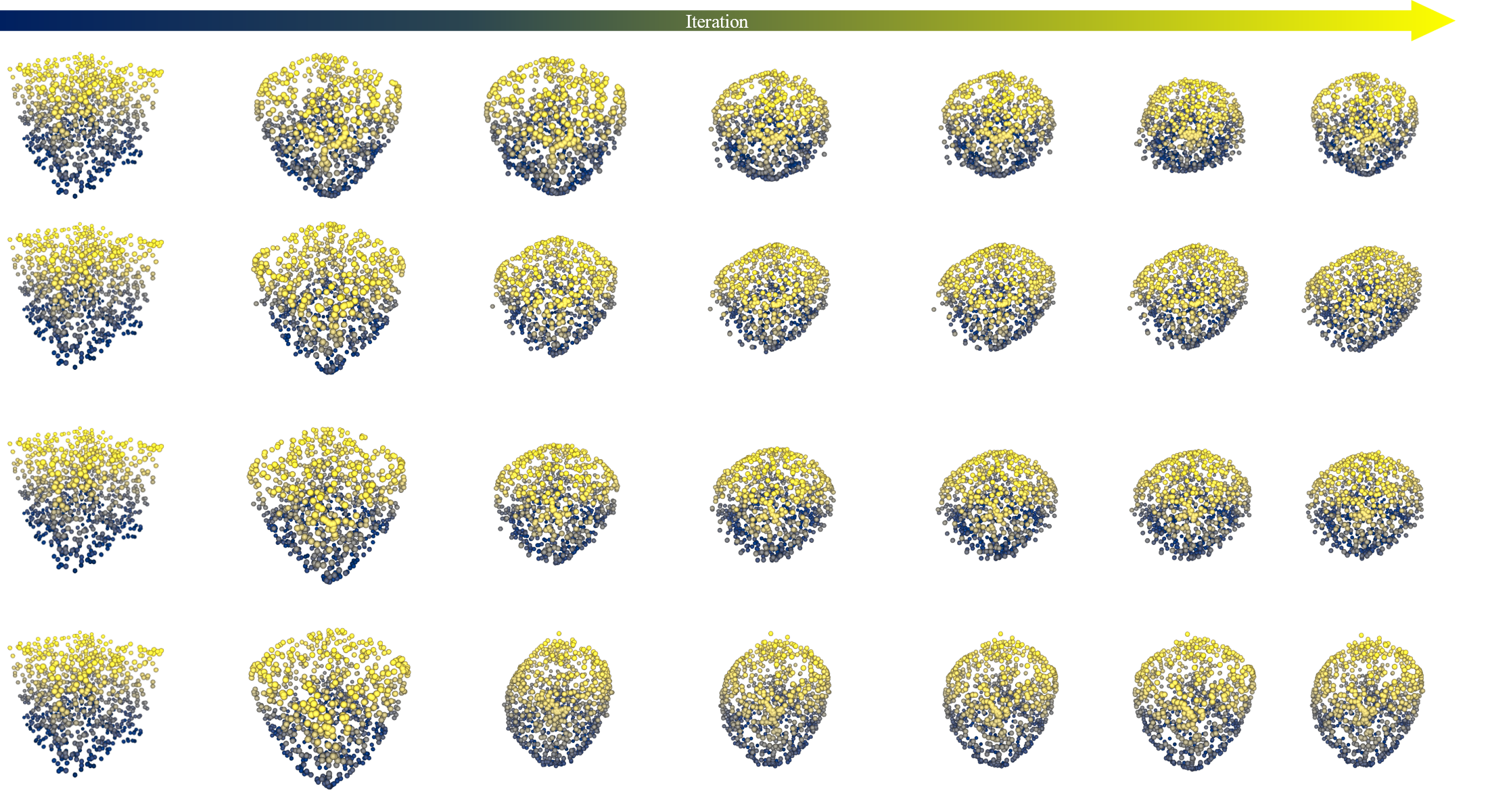}} \\
        \subfloat[]{\includegraphics[width=\textwidth]{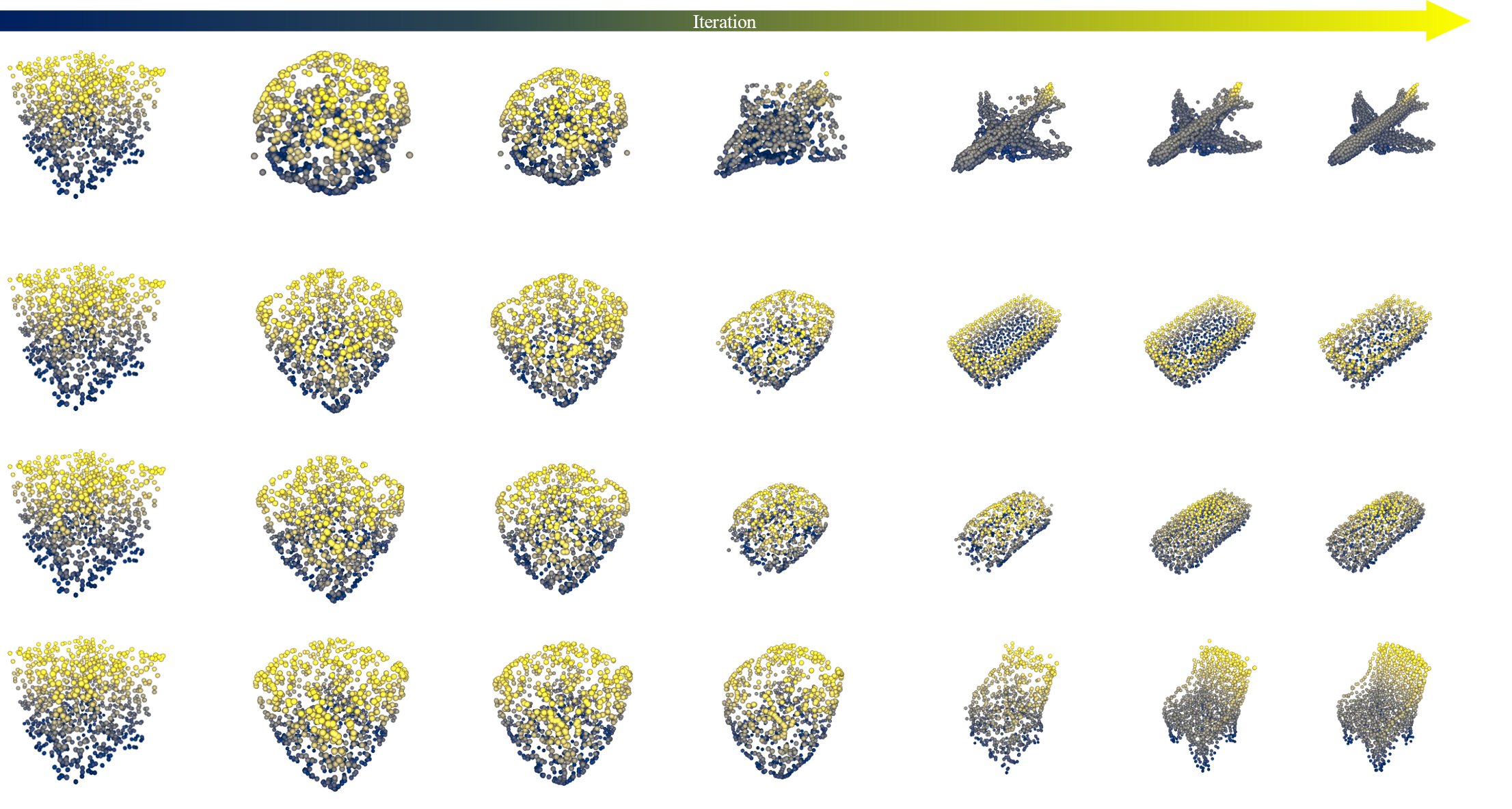}}
        \caption{Optimization process in the ModelNet40 at PPC 1 initialized from uniform noise, distilled by using (a) DM~\cite{dm} and (b) the proposed method.}
        \label{fig:supp_fig6}
\end{figure*}

Figure~\ref{fig:supp_fig6} compares the optimization process for DM~\cite{dm} and the proposed method. DM fails to converge even after multiple training iterations, however the proposed method progressively refines the point cloud models into more structured shapes.

\begin{figure*}[t]
    \centering
        \subfloat[PPC1]{
        \includegraphics[width=0.071\textwidth]{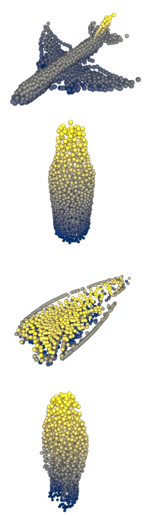}
        \label{fig:supp_fig7a} } 
        \subfloat[PPC3]{
        \includegraphics[width=0.214\textwidth]{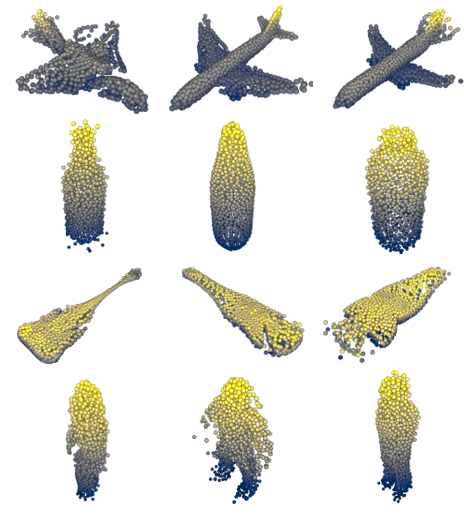}
        \label{fig:supp_fig7b} } 
        \subfloat[PPC10]{
        \includegraphics[width=0.655\textwidth]{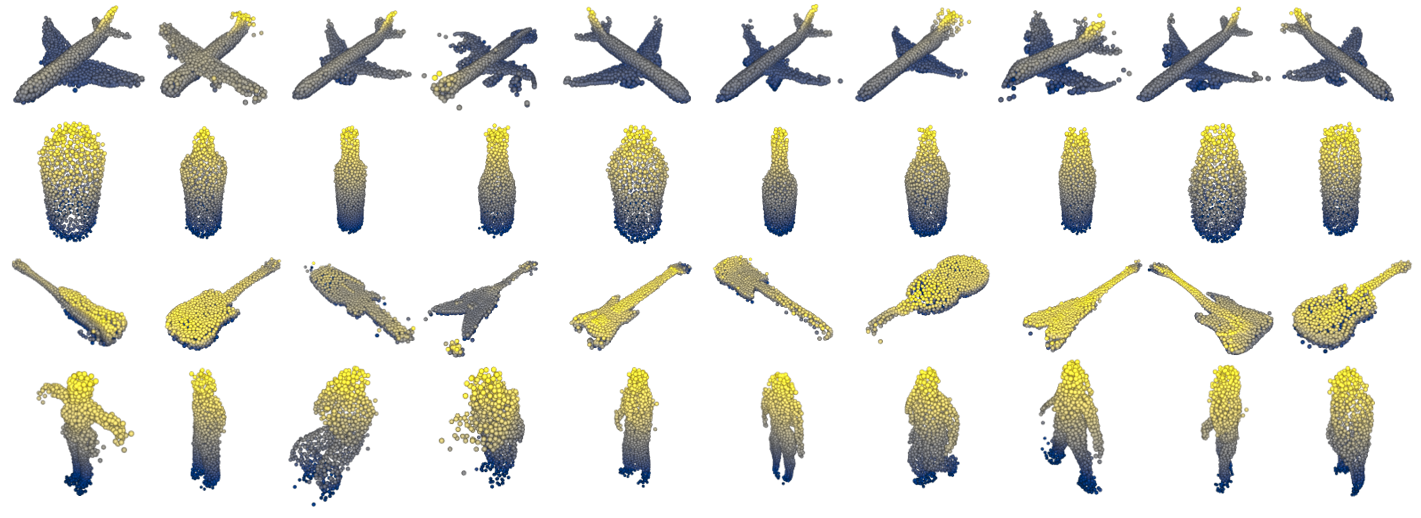}
        \label{fig:supp_fig7b} }
        \caption{The synthetic dataset distilled from ModelNet40 by using the proposed method at three PPC values of 1, 3, and 10, respectively.}
        \label{fig:supp_fig7}
\end{figure*}

\begin{figure*}[t]
	\centering
        \includegraphics[width=\textwidth]{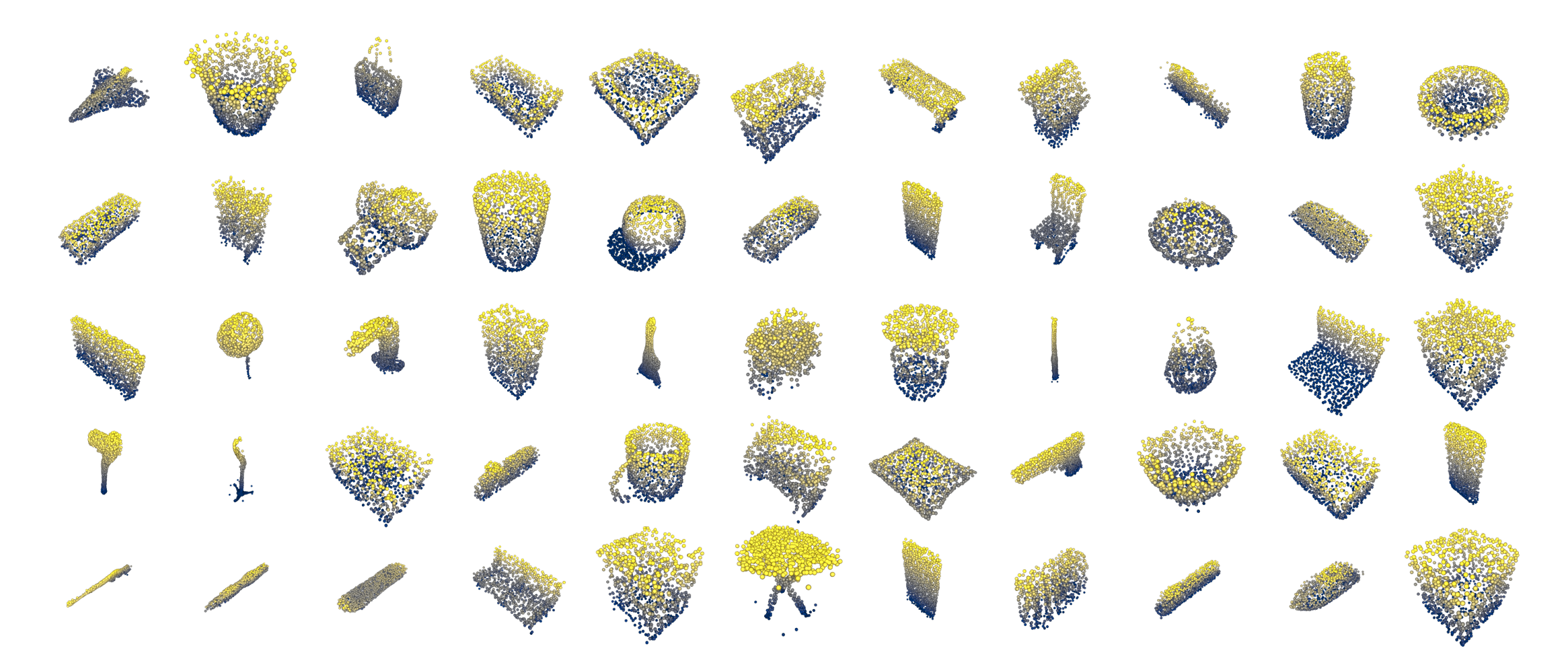}
        \caption{The synthetic dataset distilled from ShapeNet by using the proposed method at PPC 1.}
        \label{fig:supp_fig8}
\end{figure*}

\begin{figure*}[t]
	\centering
        \includegraphics[width=0.6\linewidth]{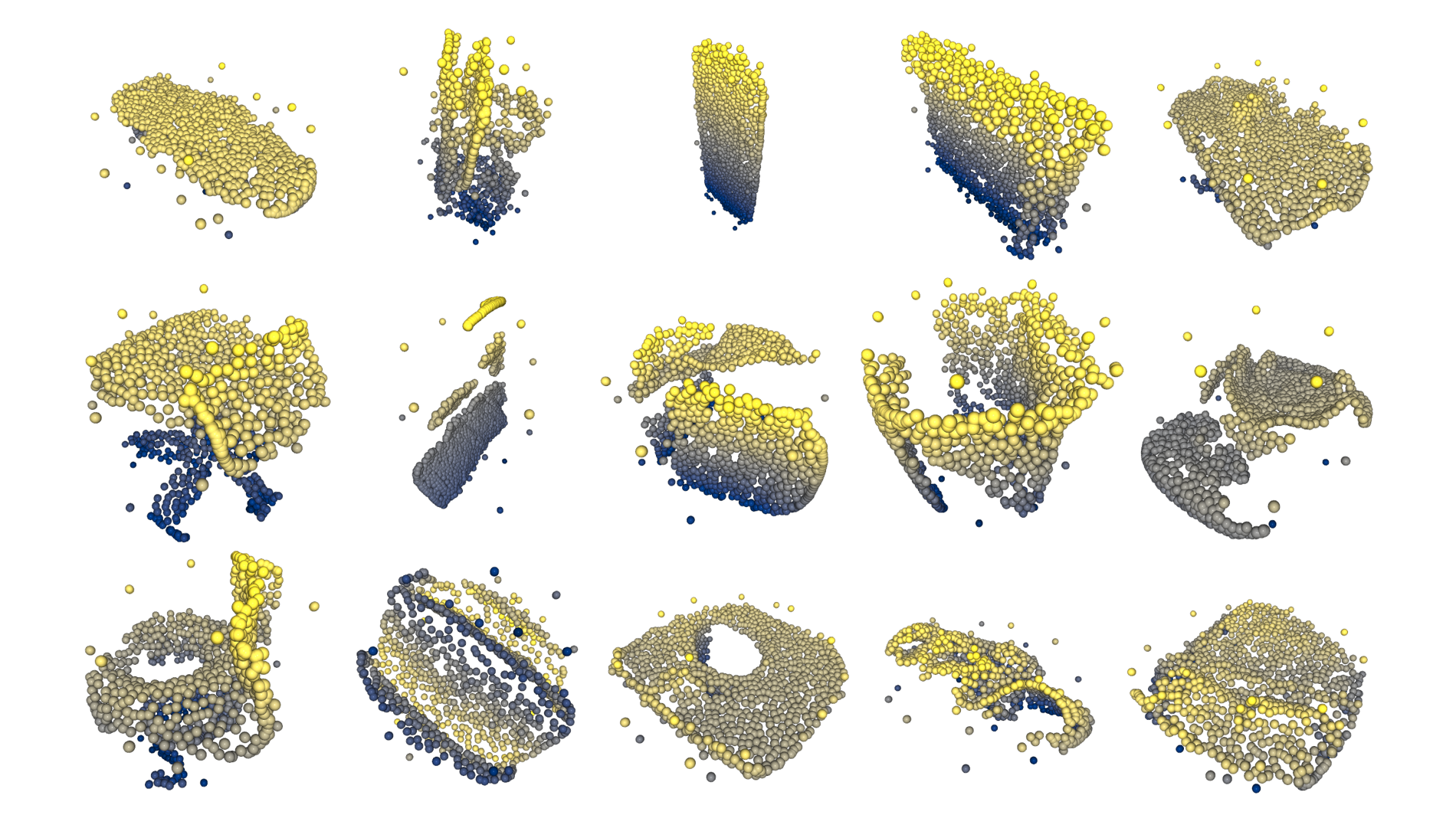}
        \caption{The synthetic dataset distilled from ScanObjectNN by using the proposed method at PPC 1.}
        \label{fig:supp_fig9}
\end{figure*}

Figure~\ref{fig:supp_fig7} shows the results of the proposed method on ModelNet40 at different PPC values: 1, 3, and 10. As PPC increases, the distilled synthetic datasets exhibit more diverse shapes and orientations. In particular, as shown in the last row, the human class appears in diverse poses.
Additional qualitative results of the proposed method on the ShapeNet and ScanObjectNN datasets can be found in Figures~\ref{fig:supp_fig8} and \ref{fig:supp_fig9}, respectively. 

\end{document}